\documentclass{article}
\usepackage{amsmath,amssymb,enumerate,theorem}

\newcommand{\nin}{\not\in}
\newcommand{\tmem}[1]{{\em #1\/}}
\newcommand{\tmop}[1]{\ensuremath{\operatorname{#1}}}
\newcommand{\tmstrong}[1]{\textbf{#1}}
\newcommand{\tmtextit}[1]{{\itshape{#1}}}
\newcommand{\tmtexttt}[1]{{\ttfamily{#1}}}
\newenvironment{enumerateroman}{\begin{enumerate}[i.] }{\end{enumerate}}
\newenvironment{itemizeminus}{\begin{itemize} }{\end{itemize}}
\newtheorem{definition}{Definition}
{\theorembodyfont{\rmfamily}\newtheorem{example}{Example}}
\newtheorem{proposition}{Proposition}
\newtheorem{theorem}{Theorem}

\begin{document}

\title{Combinatorial explorations in Su-Doku}\author{jmc}\maketitle

\begin{abstract}
  Su-Doku, a popular combinatorial puzzle, provides an excellent testbench for
  heuristic explorations. \ Several interesting questions arise from its
  deceptively simple set of rules. \ How many distinct Su-Doku grids are
  there? \ How to find a solution to a Su-Doku puzzle? Is there a unique
  solution to a given Su-Doku puzzle? \ What is a good estimation of a
  puzzle's difficulty? \ What is the minimum puzzle size (the number of
  ``givens'')?
  
  This paper explores how these questions are related to the well-known
  alldifferent constraint which emerges in a wide variety of Constraint
  Satisfaction Problems (CSP) and compares various algorithmic approaches
  based on different formulations of Su-Doku.
\end{abstract}

\section{Su-Doku as a CSP}

\paragraph{Su-Doku grids and puzzles.}Su-Doku is a well-known logic-based
number placement puzzle. \ The objective is to fill a 9x9 square grid so that
each line, each column or file, and each of the nine 3x3 blocks contains
exclusively the digits 1 to 9, only once each. \ A puzzle is a partially
completed grid.

This definition is readily generalized to larger grids. \ Let us consider $M_n
=\{1 \ldots n\}$ the set of digits 1 to $n$, a Su-Doku of size $n$ is a $n^2
\times n^2$ grid which is to be filled so that each of the $n^2$ lines, each
of the $n^2$ columns and each of the the $n^2$ blocks contains the digits 1 to
$n$ only one time each.

The deceptively simple definition actually hides large combinatorial problems,
even for small values of $n$. \ This is, for instance, reflected in the number
of such Su-Doku grids as calculated by Felgenhauer and Jarvis {\cite{FJ}}:

\begin{center}
  \begin{tabular}{ll}
    {\tmstrong{Size}} & {\tmstrong{Numb{\tmstrong{}}er}}\\
    1 & 1\\
    2 & 288\\
    3 & 6,670,903,752,021,072,936,960\\
    {\tmstrong{Table 1.}} & 
  \end{tabular}{\tmstrong{}}
\end{center}

This is sequence A107739 in the {\tmem{Online Encyclopedia of Integer
Sequences}}. The number of grids in the familiar size-3 Su-Doku is already
mind-boggling!

\paragraph{Constraint Satisfaction Problem.}The rules of Su-Doku can be cast
in terms of constraints that a solution should comply with to be valid. \ A
constraint satisfaction problem (CSP) is a triple $(X, D, C)$, where $X$ is a
sequence of $n$ variables $x_1, x_2, \ldots x_n$, $D$ is a sequence of $n$
finite domains $D_1, D_2, \ldots D_n$, where $D_i$ is the set of possible
values for variable $x_i$, and $C$ a finite set of constraints between
variables.

A constraint $C$ on a subsequence of variables $x_{i_1}, x_{i_2}, \ldots
x_{i_m}$ is simply a subset of the cartesian product $D_{i_1} \times D_{i_2}
\ldots D_{i_m}$, which expresses the allowed combination of variable values.

Constraints are often expressed as equations that variables must satisfy.

\paragraph{General CSP Expression for Su-Doku.}As will be seen in later
sections, there are several ways to express a Su-Doku puzzle as a concrete
CSP. \ In somewhat abstract terms, the CSP formulation would state four groups
of constraints:
\begin{enumerateroman}
  \item Each of the $n^4$ cells contains a unique value from the set $M_{n^2}$
  
  \item Each of the $n^2$ lines of the grid contains all values from the set
  $M_{n^2}$
  
  \item Each of the $n^2$ files of the grid contains all values from the set
  $M_{n^2}$
  
  \item Each of the $n^2$ blocks of the grid contains all values from the set
  $M_{n^2}$
\end{enumerateroman}
where each of $n^4$ variables $x_1, \ldots, x_{n^4}$ would represent a grid
cell.

The initial ``givens'' of a Su-Doku puzzle restrict the actual sets from which
the above constraints allow values to be assigned to these variables. \ In the
conventional Su-Doku puzzle, there are 81 variables and $81 + 3 \times 9 =
108$ such general constraints.

\subsection{Constraint propagation v. Search}

\paragraph{Alternating propagation and search.}When nothing much is known
about the constraints of a CSP, the general procedure to find a solution, or
to show that none exists, alternates {\tmem{propagation{\tmem{}}}} and
{\tmem{search{\tmem{}}}}. \ Each constraint is associated with a propagation
procedure which tries to shrink the domains of the constraint variables by
removing values that are certainly not part of a solution. \ This procedure
uses first the locally available information, i.e. information provided by the
constraint itself. \ When such a value is excluded from a variable's domain,
this information is also provided to all other propagation procedures
associated with constraints that share the same variable. \ This externally
provided information might then trigger a propagation procedure of a
constraint which was locally consistent, which in turns might exclude other
values, providing again this new information to other constraint procedures.
The cascade of propagation procedure calls stops, however, after some time, in
a state of global stability where all of the CSP constraints are locally
consistent together.

The globally stable state may or may not be a solution (or a failure) to the
CSP. \ If not, a search procedure splits the current problem into at least two
CSPs, usually by assigning a value to a variable from its current domain. \
More generally this is done by dividing the current $D_i$ domain of a variable
$x_i$ into $k$ disjoint subsets ($k \geqslant 2$) the union of which is $D_i$
and considering the $k$ CSP subproblems. \ The propagation phase is repeated
for each subproblem, thus recursively creating a search tree with the original
problem at its root and the solutions or failures at its leaves. \ It is
indeed a search procedure since the choice of the domain to split, and hence
of which variable to propagate, and how to split it may be based on
{\tmem{heuristics{\tmem{}}{\tmem{}}}} or rules of thumb specific to the
problem at hand.

\paragraph{Human Problem-Solving in Su-Doku puzzles.}It is generally
acknowledged that the alternating propagation and search procedures are
effectively used by human solvers when tackling Su-Doku puzzles. \ All
tutorials, books, Web sites and other broadly available Su-Doku instructional
material start by highlighting that the first steps to be taken towards a
solution involve propagation of the givens, simply by ticking off for each of
them its value from the other cells in the same line, file and block. \ Should
this leave exactly one possible value for another cell, mark it and propagate
in turn to this cell's line, file and block until no more propagation is
possible.

At this step, there are litterally tens of heuristics with inventive names
such as ``Fish'', ``XY-Wing'' or ``XY-Chain'', ranging from the simple check
to the complex pattern, to choose from in order to perform the search phase. \
Numerous Su-Doku Web sites offer catalogs of patterns to look for in a
partially solved puzzles, and each human solver develops his or her own
individual catalog as experience grows.

There is no recognized standard, however, for evaluating the complexity of
such heuristics. \ Hence there is no simple way of comparing Su-Doku puzzles
difficulties, as some search heuristics may be considered simple by some and
complex by others. \ While the number of givens, which are in fact starting
points for the propagation procedures, is certainly an indication of the
difficulty level, it is not a complete indicator of the actual complexity. \
Even so-called {\tmem{minimal puzzles{\tmem{}}}} where for $n = 3$ there are
17 givens, might prove easy to solve if the initial propagation goes far
towards the solution, leaving only few empty cells for search. (The fact that
a $n = 3$ Su-Doku puzzle needs at least 17 givens in order to have a unique
solution is still, at the time of this writing, a conjecture.)

\subsection{A Simple Propagation-Search Algorithm}

In this section we present a very simple implementation of the alternated
propagation-search phases to solve Su-Doku puzzles as CSP. \ This is for
illustrative purpose and by no means the only way to implement propagation and
search, or to strike a balance between propagation and search in CSP
solutions. \ Some of the ideas here are inspired by {\cite{Norvig}}, and, for
lack of a better name, we simply call this algorithm the PS-1-2 algorithm.

\paragraph{Propagation.}With each cell in the grid, the algorithm maintains an
array of the valid values which can be used for this cell, its so-called
{\tmem{domain}} that the propagation phase seeks to reduce as much as possible
provided the constraints.

Initially for a $n$ order Su-Doku puzzle, all domains $D_{i, j}$ are the same
set $M_{n^2}$ of the first $n^2$ integers.

Propagation resolves into iterating four separate steps:
\begin{itemizeminus}
  \item A value $v$ is assigned to a cell, thus reducing $D_{i, j}$ to $\{v\}$
  
  \item The newly assigned value $v$ is deleted from the domains of cells
  lying in the same line, file and block than the initial cell: $D_{i, k}$,
  $D_{k, j}$ and $D_{B (i, j) = B (p, q)}$ are therefore reduced to $D_{i, k}
  \backslash \{v\}$, $D_{k, j} \backslash \{v\}$ and $D_{B (i, j) = B (p, q)}
  \backslash \{v\}$ respectively. (Here $i, j$ range over $1 \ldots n^2$, and
  so do $k, p$ and $q$; the somewhat informal notation $B (i, j)$ denotes the
  block, in the range $1 \ldots n^2$, containing the $i, j$ cell.)
  
  \item Another reduction step is taken, leveraging the duality of constraints
  in the particular Su-Doku CSP formulation. \ As each line, file or block may
  only contain one occurence of any number $1 \ldots n^2$, each of the number
  has to be assigned to a unique cell within a line, file or block. So if a
  value $v$ appears only once in all $D_{i, j}$ of a given line, file or
  block, the process reduces this singled out $D_{i, j}$ to $\{v\}$.
  
  \item After this reduction step, some domains $D_{i, j}$ may happen to be
  empty, in which case the puzzle has no solution - we'll say that the
  propagation is {\tmem{blocked}} - or to be reduced to a singleton $\{w\}$.
  If all domains are singletons, the puzzle is {\tmem{solved}}. \ If only some
  of the domains are singletons, the algorithm reiterates these steps
  assigning the singletons' values to the corresponding cells.
\end{itemizeminus}
The iteration is stopped when no further reduction happens in step 4 of the
above propagation process. \ Reductions are done in any order as it does not
impact the final result after the system reaches a quiescent state. \ The
``1'' in the algorithm name comes from the choice of reducing domains on a
single constraint type (and its dual): the unicity of values for CSP
variables.

\paragraph{Data representation.}In order to lower the computation costs, the
domains for each of the $n^2$ variables representing the puzzle cells are
implemented as {\tmem{packed arrays}} in C. \ Reduction then becomes a logical
operation on a bit arrays. \ Step 3 of the previous propagation process
requires the domains to be transposed: for each line, file and block, $n^2$
new bit arrays are computed, the i-th of which is made of bits $i$ of the
$n^2$ domain bit arrays.

\paragraph{Binary Search.}After the propagation phase we may end in a blocked,
solved or yet indeterminate state.

Again we choose here a simple search procedure for the next steps towards a
solution in the indeterminate case. Namely we look for domains reduced to
simple pairs $\{v, w\}$ and we operate a binary depth-first search, first
reducing to $\{v\}$ and then, if no solution is found after recursive
propagation and search, reducing to $\{w\}$ and propagating/searching again.

Note that if at the end of each of the propagation phase we do indeed find
pair domains, this process will certainly terminate either finding a solution
or proving, after enumeration, that no solution exist for this puzzle.

The ``2'' in the name of the algorithm derives from the fact that we only
consider pair domains in the search phase.

\paragraph{}The PS-1-2 algorithm is only one in a scope of algorithms which we
will investigate further in the following sections and which were developed to
solve a particular type of constraint, called {\tmem{alldifferent}} in the CSP
literature. It so happens that the CSP formulation of Su-Doku uses only
inter-related alldifferent constraints thus offering a perfect case study for
combinatorial analyses of the various approaches to the general solution of
alldifferent constraints. Even more so, we will see that Su-Doku constraints
are a special form of alldifferent constraints, i.e {\tmem{permutation}}
constraints, which specific properties will suggest a completely different
approach to Su-Doku puzzle representation and solution explored in the second
section of this paper.

\subsection{Experiments and Results}

\paragraph{A C Implementation.}The PS-1-2 algorithm was implemented in C under
Cygwin for experimentation purposes. \ The core of the implementation is
articulated around two functions: a propagation and reduction function called
\tmtexttt{solveStep}, and a recursive depth-first search function called
\tmtexttt{pairReduce}.
\begin{verbatim}
int solveStep( int main_step ){ 
  int step, i, flag, main_flag = 1;
  while( main_flag ){ 
    // Local rules and propagation main loop 
    flag = 1; 
    step =1; 
    // 1. Propagate givens 
    while( flag ){ 
      flag = propagate(); 
      step++; 
    } 
    main_step += step; 

    // 2. Reduces lines, cols and blocks 
    // Rem.: flag is 0 at this point, L is n*n 
    for( i = 0; i < L; i++ ) flag += reduceLine( i ); 
    for( i = 0; i < L; i++ ) flag += reduceColumn( i ); 
    for( i = 0; i < L; i++ ) flag += reduceBlock( i ); 
    main_step += flag; 

    main_flag = (flag > 0 ) ? 1 : 0;
    }
  return main_step;
}
\end{verbatim}
The previous code fragment details the \tmtexttt{solveStep} function which
propagates assignments of values to cells by calling the (not-represented)
\tmtexttt{propagate} function, which in turn operates on the domain bit array
representations, deleting the assigned values from other cells' domains in
each relevant line, file and block. This is in fact step 2 of the PS-1-2
algorithm as described in the previous section. \ Then the dual step in domain
reduction is taken by calling the (not-represented) \tmtexttt{reduceLine},
\tmtexttt{reduceColumn} and \tmtexttt{reduceBlock} functions which handle the
transposition and reduction in step 3 of the PS-1-2 algorithm.

This function exits when no domain can be further reduced to a singleton
through the iteration of the basic \tmtexttt{propagate} and \tmtexttt{reduce}
operations.

In addition the function maintains various counters, namely \tmtexttt{step}
and \tmtexttt{main\_step}, for simple statistics.

The depth-first search function is straightforward:
\begin{verbatim}
int pairReduce( int step ){ 
  packPtr p; 
  struct pack keep[S];
  if( 1 == solvedp() ) return step; 
  if( 1 == blockedp() ) return step;

  S_ReduceSteps += 1;
  // Find a pair domain p 
  p = nextPair(); 
  if( (packPtr)0 == p ){ 
    printf( "No pair left. S=%d, B=%d
", solvedp(), blockedp() ); } 
  else{ 
    int hi, lo;
    // Store current state of search as an array of domain bit arrays 
    packcpy( keep, cell ); 
    // Extract low and high values in pair p 
    lo = getPack( p );
    // Delete low value from domain p
    subPack( p, lo ); 
    hi = getPack( p ); 
    // And propagate to other domains
    step = solveStep( step ); 
    if( 1 == solvedp() ) return step; 

    step = pairReduce( step ); 
    if( 1 == solvedp() ) return step;

    // No solution reached, restore search state
    packcpy( cell, keep );
    // Now delete high value from domain p
    subPack( p, hi );
    // And propagate to other domains 
    step = solveStep( step );  
    if( 1 == solvedp() ) return step; 
    if( 0 == blockedp() ){ 
      step = pairReduce( step );
    } 
  } 
  return step; 
}
\end{verbatim}
This code fragment illustrates the search procedure. If \tmtexttt{pairReduce}
is entered in a solved or blocked state it returns immediately. If entered in
an indeterminate state, it first finds a pair domain, by calling the
(non-represented) \tmtexttt{nextPair} function. (In the implementation this
function does a simple but costly linear search on the array of all domain bit
arrays.) If it fails to find such a pair domain it simply stops, although in
all of the test puzzles this never happened.

When it succeeds, however, the function backs up the current search state,
here an array of domain bit arrays representing the remaining possible values
for each cell in the puzzle, assigns first the highest value of the pair
domain to the cell and propagates this assignment by calling the previously
mentioned \tmtexttt{solveStep}. \ At this point, the puzzle is either solved
and \tmtexttt{pairReduce} returns (recursively up to the first caller in
fact), or in a blocked or indeterminate state. \ In the latter cases, and in
standard depth-first fashion, we search another pair domain by calling
recursively \tmtexttt{pairReduce}.

Note that if the state is blocked, this recursive call returns immediately.
When it does not and we still have no solution on the first branch of our
binary search, we try the other one. \ The function duly restores the backed
up state of search and this time assigns the lowest value of the pair domain
to the cell and propagates to other domains, again calling
\tmtexttt{solveStep}.

Various counters are also updated for statistical purposes. \ In order to
obtain a running solver program, these functions are wrapped into a main
function which initializes all domains to the same all-one bit arrays, reads
the puzzle in from a file, executes the initial propagation of the ``givens''
and calls \tmtexttt{pairReduce(0)}.

\paragraph{An Example Run.}We ran PS-1-2 on some of the minimal puzzle
instances as collected by Gordon Royle {\cite{Royle}} who maintains a catalog
of order 3 Su-Dokus puzzles with only 17 ``givens''.

On another example, the following puzzle, which is not part of this ``minimal
puzzles'' set:

\begin{center}
  
\end{center}

\begin{example}
  Puzzle

  \tmtexttt{
  
  .125.487.
  
  .........
  
  75.....23
  
  ..41.87..
  
  .2..5..4.
  
  ..34.95..
  
  48.....17
  
  .........
  
  .357.169.}
\end{example}

is solved with only 77 propagation and 11 search operations by PS-1-2, as
detailed in the following tabular trace:

\begin{example}

  \tmtexttt{Prop \ \ \ Red \ \ \ \ Srch \ \ \ Tot. Prop \ \ \ \ \ \ Tot. Srch
  
  3 \ \ \ \ \ \ - \ \ \ \ \ \ - \ \ \ \ \ \ 3 \ \ \ \ \ \ \ \ \ \ \ \ \ \ 0
  
  - \ \ \ \ \ \ 3 \ \ \ \ \ \ - \ \ \ \ \ \ 6 \ \ \ \ \ \ \ \ \ \ \ \ \ \ 0
  
  3 \ \ \ \ \ \ - \ \ \ \ \ \ - \ \ \ \ \ \ 9 \ \ \ \ \ \ \ \ \ \ \ \ \ \ 0
  
  - \ \ \ \ \ \ 0 \ \ \ \ \ \ - \ \ \ \ \ \ 9 \ \ \ \ \ \ \ \ \ \ \ \ \ \ 0
  
  - \ \ \ \ \ \ - \ \ \ \ \ \ h \ \ \ \ \ \ 9 \ \ \ \ \ \ \ \ \ \ \ \ \ \ 1
  
  3 \ \ \ \ \ \ - \ \ \ \ \ \ - \ \ \ \ \ \ 12 \ \ \ \ \ \ \ \ \ \ \ \ \ 1
  
  - \ \ \ \ \ \ 0 \ \ \ \ \ \ - \ \ \ \ \ \ 12 \ \ \ \ \ \ \ \ \ \ \ \ \ 1
  
  - \ \ \ \ \ \ - \ \ \ \ \ \ h \ \ \ \ \ \ 12 \ \ \ \ \ \ \ \ \ \ \ \ \ 2
  
  3 \ \ \ \ \ \ - \ \ \ \ \ \ - \ \ \ \ \ \ 15 \ \ \ \ \ \ \ \ \ \ \ \ \ 2
  
  - \ \ \ \ \ \ 3 \ \ \ \ \ \ - \ \ \ \ \ \ 18 \ \ \ \ \ \ \ \ \ \ \ \ \ 2
  
  3 \ \ \ \ \ \ - \ \ \ \ \ \ - \ \ \ \ \ \ 21 \ \ \ \ \ \ \ \ \ \ \ \ \ 2
  
  - \ \ \ \ \ \ 2 \ \ \ \ \ \ - \ \ \ \ \ \ 23 \ \ \ \ \ \ \ \ \ \ \ \ \ 2
  
  3 \ \ \ \ \ \ - \ \ \ \ \ \ - \ \ \ \ \ \ 26 \ \ \ \ \ \ \ \ \ \ \ \ \ 2
  
  - \ \ \ \ \ \ 0 \ \ \ \ \ \ - \ \ \ \ \ \ 26 \ \ \ \ \ \ \ \ \ \ \ \ \ 2
  
  - \ \ \ \ \ \ - \ \ \ \ \ \ h \ \ \ \ \ \ 26 \ \ \ \ \ \ \ \ \ \ \ \ \ 3
  
  3 \ \ \ \ \ \ - \ \ \ \ \ \ - \ \ \ \ \ \ 29 \ \ \ \ \ \ \ \ \ \ \ \ \ 3
  
  - \ \ \ \ \ \ 0 \ \ \ \ \ \ - \ \ \ \ \ \ 29 \ \ \ \ \ \ \ \ \ \ \ \ \ 3
  
  - \ \ \ \ \ \ - \ \ \ \ \ \ h \ \ \ \ \ \ 29 \ \ \ \ \ \ \ \ \ \ \ \ \ 4
  
  5 \ \ \ \ \ \ - \ \ \ \ \ \ - \ \ \ \ \ \ 34 \ \ \ \ \ \ \ \ \ \ \ \ \ 4
  
  - \ \ \ \ \ \ 1 \ \ \ \ \ \ - \ \ \ \ \ \ 35 \ \ \ \ \ \ \ \ \ \ \ \ \ 4
  
  3 \ \ \ \ \ \ - \ \ \ \ \ \ - \ \ \ \ \ \ 38 \ \ \ \ \ \ \ \ \ \ \ \ \ 4
  
  - \ \ \ \ \ \ 0 \ \ \ \ \ \ - \ \ \ \ \ \ 38 \ \ \ \ \ \ \ \ \ \ \ \ \ 4
  
  - \ \ \ \ \ \ - \ \ \ \ \ \ h \ \ \ \ \ \ 38 \ \ \ \ \ \ \ \ \ \ \ \ \ 5
  
  3 \ \ \ \ \ \ - \ \ \ \ \ \ - \ \ \ \ \ \ 41 \ \ \ \ \ \ \ \ \ \ \ \ \ 5
  
  - \ \ \ \ \ \ 0 \ \ \ \ \ \ - \ \ \ \ \ \ 41 \ \ \ \ \ \ \ \ \ \ \ \ \ 5
  
  - \ \ \ \ \ \ - \ \ \ \ \ \ h \ \ \ \ \ \ 41 \ \ \ \ \ \ \ \ \ \ \ \ \ 6
  
  5 \ \ \ \ \ \ - \ \ \ \ \ \ - \ \ \ \ \ \ 46 \ \ \ \ \ \ \ \ \ \ \ \ \ 6
  
  - \ \ \ \ \ \ 1 \ \ \ \ \ \ - \ \ \ \ \ \ 47 \ \ \ \ \ \ \ \ \ \ \ \ \ 6
  
  3 \ \ \ \ \ \ - \ \ \ \ \ \ - \ \ \ \ \ \ 50 \ \ \ \ \ \ \ \ \ \ \ \ \ 6
  
  - \ \ \ \ \ \ 0 \ \ \ \ \ \ - \ \ \ \ \ \ 50 \ \ \ \ \ \ \ \ \ \ \ \ \ 6
  
  - \ \ \ \ \ \ - \ \ \ \ \ \ l \ \ \ \ \ \ 50 \ \ \ \ \ \ \ \ \ \ \ \ \ 6
  
  3 \ \ \ \ \ \ - \ \ \ \ \ \ - \ \ \ \ \ \ 53 \ \ \ \ \ \ \ \ \ \ \ \ \ 6
  
  - \ \ \ \ \ \ 0 \ \ \ \ \ \ - \ \ \ \ \ \ 53 \ \ \ \ \ \ \ \ \ \ \ \ \ 6
  
  - \ \ \ \ \ \ - \ \ \ \ \ \ h \ \ \ \ \ \ 53 \ \ \ \ \ \ \ \ \ \ \ \ \ 7
  
  3 \ \ \ \ \ \ - \ \ \ \ \ \ - \ \ \ \ \ \ 56 \ \ \ \ \ \ \ \ \ \ \ \ \ 7
  
  - \ \ \ \ \ \ 1 \ \ \ \ \ \ - \ \ \ \ \ \ 57 \ \ \ \ \ \ \ \ \ \ \ \ \ 7
  
  3 \ \ \ \ \ \ - \ \ \ \ \ \ - \ \ \ \ \ \ 60 \ \ \ \ \ \ \ \ \ \ \ \ \ 7
  
  - \ \ \ \ \ \ 0 \ \ \ \ \ \ - \ \ \ \ \ \ 60 \ \ \ \ \ \ \ \ \ \ \ \ \ 7
  
  - \ \ \ \ \ \ - \ \ \ \ \ \ h \ \ \ \ \ \ 60 \ \ \ \ \ \ \ \ \ \ \ \ \ 8
  
  5 \ \ \ \ \ \ - \ \ \ \ \ \ - \ \ \ \ \ \ 65 \ \ \ \ \ \ \ \ \ \ \ \ \ 8
  
  - \ \ \ \ \ \ 0 \ \ \ \ \ \ - \ \ \ \ \ \ 65 \ \ \ \ \ \ \ \ \ \ \ \ \ 8
  
  - \ \ \ \ \ \ - \ \ \ \ \ \ h \ \ \ \ \ \ 65 \ \ \ \ \ \ \ \ \ \ \ \ \ 9
  
  3 \ \ \ \ \ \ - \ \ \ \ \ \ - \ \ \ \ \ \ 68 \ \ \ \ \ \ \ \ \ \ \ \ \ 9
  
  - \ \ \ \ \ \ 0 \ \ \ \ \ \ - \ \ \ \ \ \ 68 \ \ \ \ \ \ \ \ \ \ \ \ \ 9
  
  - \ \ \ \ \ \ - \ \ \ \ \ \ h \ \ \ \ \ \ 68 \ \ \ \ \ \ \ \ \ \ \ \ \ 10
  
  4 \ \ \ \ \ \ - \ \ \ \ \ \ - \ \ \ \ \ \ 72 \ \ \ \ \ \ \ \ \ \ \ \ \ 10
  
  - \ \ \ \ \ \ 0 \ \ \ \ \ \ - \ \ \ \ \ \ 72 \ \ \ \ \ \ \ \ \ \ \ \ \ 10
  
  - \ \ \ \ \ \ - \ \ \ \ \ \ h \ \ \ \ \ \ 72 \ \ \ \ \ \ \ \ \ \ \ \ \ 11
  
  5 \ \ \ \ \ \ - \ \ \ \ \ \ - \ \ \ \ \ \ 77 \ \ \ \ \ \ \ \ \ \ \ \ \ 11
  
  - \ \ \ \ \ \ 0 \ \ \ \ \ \ - \ \ \ \ \ \ 77 \ \ \ \ \ \ \ \ \ \ \ \ \ 11
  
  Grid: solved 1, blocked 0; in 77 operations

  -------------------------
  
  | 6 1 2 | 5 3 4 | 8 7 9 |
  
  | 3 4 9 | 2 8 7 | 1 6 5 |
  
  | 7 5 8 | 9 6 1 | 4 2 3 |
  
  -------------------------
  
  | 5 9 4 | 1 2 8 | 7 3 6 |
  
  | 8 2 7 | 6 5 3 | 9 4 1 |
  
  | 1 6 3 | 4 7 9 | 5 8 2 |
  
  -------------------------
  
  | 4 8 6 | 3 9 5 | 2 1 7 |
  
  | 9 3 1 | 7 4 2 | 6 5 8 |
  
  | 2 7 5 | 8 1 6 | 3 9 4 |
  
  -------------------------
  
  }
\end{example}

The process called the search procedure 11 times, when the
propagation/reduction operations reach quiescence as indicated by a 0 in the
Red(uctions) column. \ The Srch column indicates whether the h(igh) or l(ow)
value of the pair searched is used for the next propagation phase. \ In the
particular instance, backtrack occurred only once at the sixth pair search:
both high and low value were propagated to find the solution.

\paragraph{Conclusions.}The canonical procedure to solve CSP-formulated
problems alternates a propagation phase, where data is used to reduce domains
of the variables as far as possible, also known as filtering, with a search
phase, a backtrack procedure which explores incremental steps towards a
solution. \ There is ample room for variability in this framework both in the
balance between propagation and search, and within each phase in the criteria
used in filtering and in search.

In the case of Su-Doku puzzles, we have presented a naive algorithm, PS-1-2,
which only filters on unicity of the variable value and of this value per
group (line, file or block) in the propagation phase, and only uses binary
search in the alternating search phase. \ Although there should be
pathological cases where the binary search phase might fail, the PS-1-2
algorithm was successful at solving quickly all the puzzles we submitted,
including so-called minimal puzzles.

\subsection{CSP/SAT/LP formulations: the alldifferent constraint}

While several ad hoc CSP solving procedures may be designed for Su-Doku
puzzles, its constraints generally fall under a now well documented class of
constraints for which efficient filtering procedures have been published in
the literature and are embedded in several tools, commercial and otherwise. \
The pattern appearing in all the constraints in the above CSP formulation of
Su-Doku directly relates to one of the latter, the {\tmem{alldifferent}}
constraint {\cite{VanHoeve}}.

\paragraph{A CSP alldifferent expression.}We will rephrase the CSP expression
in terms of this well studied alldifferent constraint. Let us consider the
$n$-sized Su-Doku puzzle and introduce the $n^4$ variables $x_{1, 1} \ldots
x_{n^2, n^2}$ representing the cells in a grid where, by convention, $x_{i,
j}$, is the value to be assigned to the cell in line $i$ and file $j$. All
variables have the same domain, taking their values in $M_{n^2}$. The CSP
expression of the problem is to find a unique value for each variable
satisfying the following set of alldifferent constraints:
\begin{itemize}
  \item $\forall i \in \{1, \ldots n^2 \} \tmop{alldifferent} (x_{1, i},
  \ldots x_{n^2, i})$
  
  \item $\forall j \in \{1, \ldots n^2 \} \tmop{alldifferent} (x_{j, 1},
  \ldots x_{j, n^2})$
  
  \item $\forall b \in \{1, \ldots n^2 \} \tmop{alldifferent} (x_{b_1}, \ldots
  x_{b_{n^2}})$ where the $b_k$ are the pair of indices of variables
  representing cells in the same block
\end{itemize}
\paragraph{SAT formulations.}A given alldifferent constraint naturally
translates into a set of simpler binary constraints on its variables, the
{\tmem{naive}} translation. In such naive translation the alldifferent
constraint is expressed as a conjunction of disjunctive clauses involving at
most two variables and the values of the variable domains.

For instance in the size 2 Su-Doku puzzles, an alldifferent constraint on four
variables as in line, file and block constraints becomes:

alldifferent($x_1, x_2, x_3, x_4$) $\rightleftharpoons$

$\hspace*{\fill} x_1 \neq x_2 \wedge x_1 \neq x_3 \wedge x_1 \neq x_4 \wedge
x_2 \neq x_3 \wedge x_2 \neq x_4 \wedge x_3 \neq x_4 \wedge$ 
$\hspace*{\fill} (x_1 = 1 \vee x_1 = 2 \vee x_1 = 3 \vee x_1 = 4) \wedge$
$\hspace*{\fill} (x_2 = 1 \vee x_2 = 2 \vee x_2 = 3 \vee x_2 = 4) \wedge$
$\hspace*{\fill} (x_3 = 1 \vee x_3 = 2 \vee x_3 = 3 \vee x_3 = 4) \wedge$
$\hspace*{\fill} (x_4 = 1 \vee x_4 = 2 \vee x_4 = 3 \vee x_4 = 4)$

a CNF formula with 10 disjunctive clauses, 6 of which are binary and 4 of
which unary. \ Generally speaking the size $n$ alldifferent constraint naively
translates to a CNF formula with $n (n + 1) / 2$ disjunctive clauses.

The naive translation is actually enough to be fed to standard SAT solvers
such as {\tmem{maxsatz}} {\cite{maxsatz}} and {\tmem{minisat}}
{\cite{minisat}}, for instance.

\paragraph{LP formulations.}Logic programming tools can also directly use the
above SAT expressions. \ In this section we investigate the use of the
CLP(FD), or Constraint Logic Programming for Finite Domains, extension to the
GNU-Prolog implementation {\cite{gnuprolog}} in Su-Doku puzzle experiments.

Expanding on the above analysis, a naive implementation of a single
alldifferent constraint simply translates it into a corresponding set of
binary constraints, each one of which stating that a given variable is
different from the other. \ Focusing on 4x4 Su-Doku grids, i.e. $n = 2$, for
instance, such a naive implementation of the single alldifferent constraint on
four variables would then be as follows:

\begin{verbatim}
naive_all_different(X,Y,Z,Z0) :- 
X \= Y, X \= Z, X \= Z0, Y \= Z, Y \= Z0, Z \= Z0.
\end{verbatim}

which states that each variable should have a distinct value from the other
three variables in the group. \ In order to complete the size 2 Su-Doku grid
enumeration, a definition predicate, \tmtexttt{assign}, is created to define
the (unique) domain of all variables:

\begin{verbatim}
assign(1).
assign(2).
assign(3).
assign(4).
\end{verbatim}

It is a predicate which is true for each of the four admissible values of the
cells in a size 2 Su-Doku puzzle. \ To complete the GNU-Prolog program, we use
these two predicates to express all the constraints of the 4x4 grid:

\begin{verbatim}
naive_puzzle(   A00, A01, A10, A11, B00, B01, B10, B11, 
C00, C01, C10, C11, D00, D01, D10, D11 ) :-
system_time(T0),
cpu_time(T10),
real_time(T20),
assign(  A00 ),
assign(  A01 )
assign(  A10 )
assign(  A11 )
assign(  B00 )
assign(  B01 )
assign(  B10 )
assign(  B11 )
assign(  C00 )
assign(  C01 )
assign(  C10 )
assign(  C11 )
assign(  D00 )
assign(  D01 )
assign(  D10 )
assign(  D11 )

naive_all_different(A00, A01, A10, A11)
naive_all_different(B00, B01, B10, B11)
naive_all_different(C00, C01, C10, C11)
naive_all_different(D00, D01, D10, D11)

naive_all_different(A00, A01, B00, B01 )
naive_all_different(A10, A11, B10, B11 )
naive_all_different(C00, C01, D00, D01 )
naive_all_different(C10, C11, D10, D11 )

naive_all_different(A00, A10, C00, C10 )
naive_all_different(A01, A11, C01, C11 )
naive_all_different(B00, B10, D00, D10 )
naive_all_different(B01, B11, D01, D11 )
system_time(T),
cpu_time(T1),
real_time(T2)
write( 'time T0: '), write(T0), write(', time  T: ' ),write(T), nl,
write( 'time T0: '), write(T10), write(', time  T1: ' ),write(T1), nl,
write( 'time T0: '), write(T20), write(', time  T2: ' ),write(T2), nl.
\end{verbatim}

The predicates assigns unique values to the 16 variables representing the
corresponding cells of the Su-Doku grid or puzzle, and expresses the
alldifferent constraint on each of the 4 lines, files and blocks. \ It also
keeps track of various execution times for instrumentation purposes.

Execution times are unsurprisingly very long, even for the size 2 Su-Doku
grids as the naive implementation only uses extremely local constraints within
one line, file or block and ignores global constraints. \

Fortunately, GNU-Prolog bundles a constraint logic programming extension for
finite domains which incorporates the latest filtering algorithms for a wide
variety of constraints used in industry problems. \ The CLP(FD) extension uses
specific filtering techniques for the alldifferent constraint, the theoretical
basis of which will be explored in the next sections, leading to a much more
efficient implementation of enumeration and solving tasks.

More specifically, the CLP(FD) extension offers a simple set of built-in
predicates such as: \tmtexttt{fd\_domain} to define variables' domains,
\tmtexttt{fd\_all\_different} to state a single alldifferent constraint (of
any arity), and \tmtexttt{fd\_labeling} to trigger search according to a
choice of filtering methods. \ The CLP(FD) is fully described in Diaz's Thesis
{\cite{clpfd}}. \ The size 2 Su-Doku implementation now becomes:

\begin{verbatim}
puzzle( A00, A01, A10, A11, B00, B01, B10, B11, 
C00, C01, C10, C11, D00, D01, D10, D11) :-
fd_domain(  A00, 1, 4 )
fd_domain(  A01, 1, 4 )
fd_domain(  A10, 1, 4 )
fd_domain(  A11, 1, 4 )
fd_domain(  B00, 1, 4 )
fd_domain(  B01, 1, 4 )
fd_domain(  B10, 1, 4 )
fd_domain(  B11, 1, 4 )
fd_domain(  C00, 1, 4 )
fd_domain(  C01, 1, 4 )
fd_domain(  C10, 1, 4 )
fd_domain(  C11, 1, 4 )
fd_domain(  D00, 1, 4 )
fd_domain(  D01, 1, 4 )
fd_domain(  D10, 1, 4 )
fd_domain(  D11, 1, 4 )

fd_all_different([A00, A01, A10, A11])
fd_all_different([B00, B01, B10, B11])
fd_all_different([C00, C01, C10, C11])
fd_all_different([D00, D01, D10, D11])

fd_all_different([A00, A01, B00, B01 ])
fd_all_different([A10, A11, B10, B11 ])
fd_all_different([C00, C01, D00, D01 ])
fd_all_different([C10, C11, D10, D11 ])

fd_all_different([A00, A10, C00, C10 ])
fd_all_different([A01, A11, C01, C11 ])
fd_all_different([B00, B10, D00, D10 ])
fd_all_different([B01, B11, D01, D11 ])
system_time(T0),
cpu_time(T10),
real_time(T20)
fd_labeling([A00, A01, A10, A11
             B00, B01, B10, B11
             C00, C01, C10, C11
             D00, D01, D10, D11],[variable_method(most_constrained)]),
system_time(T),
cpu_time(T1),
real_time(T2)
write( 'time T0: '), write(T0), write(', time  T: ' ),write(T), nl,
write( 'time T0: '), write(T10), write(', time  T1: ' ),write(T1), nl
write( 'time T0: '), write(T20), write(', time  T2: ' ),write(T2), nl.
\end{verbatim}

Basically this new puzzle predicate defines the 16 domains for the 16
variables with values in the same $M_4$ set, states the three group of
alldifferent constraints and finally searches for a proper labeling. \ As
before, several ancillary predicates have been added for instrumentation
purposes.

On the simple enumeration task, the first solution is almost instantly
computed on a standard Intel machine running Windows XP:

\begin{verbatim}
puzzle(    A00, A01, A10, A11, B00, B01, B10, B11, 
C00, C01, C10, C11, D00, D01, D10, D11).
time T0: 296, time  T: 296
time T0: 1609, time  T1: 1609
time T0: 155875, time  T2: 155875

A00 = 1
A01 = 2
A10 = 3
A11 = 4
B00 = 3
B01 = 4
B10 = 1
B11 = 2
C00 = 2
C01 = 1
C10 = 4
C11 = 3
D00 = 4
D01 = 3
D10 = 2
D11 = 1 ? ;
time T0: 296, time  T: 312
time T0: 1609, time  T1: 1625
time T0: 155875, time  T2: 158472

A00 = 1
A01 = 2
A10 = 3
A11 = 4
B00 = 3
B01 = 4
B10 = 1
B11 = 2
C00 = 2
C01 = 3
C10 = 4
C11 = 1
D00 = 4
D01 = 1
D10 = 2
D11 = 3 ? ;
time T0: 296, time  T: 328
time T0: 1609, time  T1: 1641
time T0: 155875, time  T2: 222624

A00 = 1
A01 = 2
A10 = 3
A11 = 4
B00 = 3
B01 = 4
B10 = 1
B11 = 2
C00 = 4
C01 = 1
C10 = 2
C11 = 3
D00 = 2
D01 = 3
D10 = 4
D11 = 1 ? ;
time T0: 296, time  T: 343
time T0: 1609, time  T1: 1656
time T0: 155875, time  T2: 228535
\end{verbatim}

and the other 288 solutions are quickly printed out for a total execution time
of 2,797 milliseconds.

\subsection{Arc-Consistency and value graph}

In order to understand how to make the best use of the local information
provided by the constraints themselves, we introduce some definitions of
{\tmem{local consistency{\tmem{.}}}}

\begin{definition}
  A constraint of arity m on the variables $x_{i_1} \ldots x_{i_m} $is
  hyperarc consistent if all values of the variables are used in some solution
  to the constraint, i.e. $\forall x_{i_k} \forall v \in D_{i_k}, \exists
  (v_{i_1}, \ldots, v_{i_{k - 1}}, v_{i_{k + 1}}, \ldots, v_{i_m}) \in D_{i_1}
  \times \ldots D_{i_{k - 1}} \times D_{i_{k + 1}} \times \ldots D_{i_m}$ such
  that $(v_{i_1}, \ldots, v_{i_{k - 1}}, v, v_{i_{k + 1}}, \ldots, v_{i_m})$
  is a solution to the constraint.
\end{definition}

\begin{definition}
  A constraint C is arc-consistent when C is of arity 2 (binary) and hyperarc
  consistent.
\end{definition}

A constraint has a solution if it can be made hyperarc consistent. \ Hyperarc
consistency is the best possible pruning based on the local information
provided by the constraint. \ The naive implementation above translate the
alldifferent constraint into a collection of binary constraints which are made
arc-consistent by filtering the domains following the simple procedure which
as soon as a domain is reduced to one value, removes this value from the
domains of all other variables.

The more efficient variants used, for instance, in the CLP(FD) package rely on
a completely different approach based on results from graph theory. \ The
correspondence with graph theory was used by R\'egin {\cite{Regin}} to create
a filtering algorithm from {\tmem{matching theory{\tmem{.}}}} \ We introduce
the notion of bipartite graph.

\begin{definition}
  Bipartite Graph. \ A graph G consists of a finite, non-empty set of elements
  V called nodes, or vertices, and a set of unordered pair of nodes E called
  edges. \ If V can be partitioned into two disjoint, non-empty sets X and Y
  such that all edges in E join a node in X to a node in Y, G is called
  bipartite with partition (X,Y); we also write G = (X,Y,E).
\end{definition}

The definition directly applies to the alldifferent constraint say, on
variable set $X =\{x_1, \ldots x_n$\} and domains $D_1 \ldots D_n$, in that it
specifies its {\tmem{value graph{\tmem{.}}}}

\begin{definition}
  Value Graph. Given an alldifferent constraint, the bipartite graph $G = (X,
  \bigcup D_i, E)$ where $(x_i, d) \in E$ iff $d \in D_i$ is called the value
  graph of the constraint.
\end{definition}

The value graph has an edge from each variable in the constraint to each of
its domain value. \ Solving such a constraint becomes a problem of
{\tmem{maximum matching}} in the corresponding value graph.

\begin{definition}
  Maximum Matching. A subset of edges in a graph G is a matching if no two
  edges have a vertex in common. \ A matching of maximum cardinality is called
  a maximum matching. \ A matching covers a set of vertices X isf every node
  in X is an endpoint of an edge in the matching.
\end{definition}

The link between matching theory and hyperarc consistency established by
R\'egin is as follows.

\begin{proposition}
  The constraint alldifferent on variable set X is hyperarc consistent iff
  every edge in its value graph belongs to a matching that covers X in the
  value graph.
\end{proposition}

Hyperarc and arc-consistency algorithms are around $O (d n^{1.5})$ where $d$
is the maximum cardinality of the domains and $n$ the number of variables.

\paragraph{Matching Theory.}Obviously if there is a complete matching from $X$
to $Y$ in $(X, Y, E)$ then for every $S \subset X$ there are at least $|S|$
vertices of $Y$ adjacent to a vertex in $S$. \ That this necessary condition
is also sufficient is usually called Halls' theorem. \ This fundamental result
was proved by Hall in 1935, but an equivalent form of it had been proved by
K\"onig and Egervary in 1931; both versions, however, follow from Menger's
theorem from 1927. \ We refer to Bollobas for demonstrations and historical
remarks {\cite{Bollobas}}.

These results will also be the origin of yet another approach to solving
Su-Doku puzzles and enumerating grids, as a complete matching is also called a
set of distinct representatives, from which are derived new expressions of the
Su-Doku problems in terms of {\tmem{exact covers}} or dually {\tmem{exact
hitting set}} problems. \ This new formulation will suggest a different
algorithmic approach which is explored in Section 2.

R\'egin's algorithm relies on the fact that if we know only one arbitrary
maximum matching, we can efficiently compute if an edge of the value graph
belongs to some matching of the same maximum size without having to explore
all such matchings.

\subsection{Bounds- and range-consistency filtering}

Hall's theorem from matching theory may also be used in relation to weaker
forms of consistency called bounds-consistency and range-consistency.

\begin{definition}
  Bounds Consistency. \ A constraint of arity m where no domain $D_i$ is
  empty, is called bounds consistent iff for each variable and each value in
  the range bounded by $\min (D_i)$ and $\max (D_i$), there exist values in
  the respective ranges bounded by the other domains minimum and maximum
  values such that together with the latter $D_i$ value they constitute a
  solution to the constraint.
\end{definition}

\begin{definition}
  Range Consistency. A constraint of arity m where no domain $D_i$ is empty,
  is called bounds consistent iff for each variable and each value in $D_i$,
  there exist values in the respective ranges bounded by the other domains
  minimum and maximum values such that together with the latter $D_i$ value
  they constitute a solution to the constraint.
\end{definition}

Note that here all domains are supposed to be integer domains which can be
(totally) ordered. \ In contrast to hyperarc and arc-consistency, bounds and
range consistency look for values in the intervals defined by domains, rather
than the domains themselves. \ As these intervals may be larger than the
actual domains, these notions represent weaker form of consistencies than
hyperarc and arc-consistency. \ Both may be considered as a
{\tmem{relaxation}} of the hyperarc consistency. (In addition, bounds
consistency may be regarded as a relaxation of range consistency itself.)

Hall's theorem has been applied by Puget to create a bounds consistency
algorithm for the alldifferent constraint {\cite{Puget}}. Given an interval
$I$ let us denote $K_I$ the set of variables $x_j$ such that $D_j \subset I$,
i.e. the subset of variables which domains are included into the considered
interval. We say that $I$ is a Hall interval iff $|I| = |K_I |$. Puget's
result is as follows.

\begin{proposition}
  The constraint alldifferent on variables $x_1, \ldots, x_n$ where no domain
  $D_i$ is empty is bounds consistent iff
  \begin{enumerateroman}
    \item for each interval $I$ $|K_I | \leq |I|$,
    
    \item for each Hall interval $I$, $\{\min (D_i), \max (D_i)\} \cap I =
    \emptyset$ for all $x_i \nin K_I$.
  \end{enumerateroman}
\end{proposition}

This can be used to create an algorithm for bounds consistency on alldifferent
constraints. \ We check every interval $I$ with bounds ranging from the
minimum of all domains to the maximum of all domains. When $|I| \leq |K_I |$
the constraint is inconsistent. \ And for each Hall interval, we remove the
minimum and/or maximum until the intersection with $I$ is empty.

\begin{example}
  Consider the following constraint:
  \begin{enumerate}
    \item alldifferent($x_1, x_2, x_3$)
    
    \item $D_1 =\{1, 2\}, D_2 =\{1, 2\}, \tmop{and} D_3 =\{2, 3\}$
  \end{enumerate}
  The intervals we need to check are $[1, 2], [1, 3], \tmop{and} [2, 3]$. \
  When $I = [1, 2]$, the domains included in the interval are $D_1$ and $D_2$,
  hence $K_I =\{x_1, x_2 \}$ and since $|I| = |K_I | = 2$, $I$ is a Hall
  interval. We only have one variable not in $K_I$, namely $x_3$, for which
  $\{\min (D_3), \max (D_3)\} \cap I =\{2\}$. \ The algorithm removes then 2
  from $D_3$ and the resulting system of sets $\{1, 2\}, \{1, 2\}, \{3\}$ is
  now bounds consistent. \ The two solutions $(1, 2, 3)$ and (2,1,3) are now a
  simple consequence of the reduction of $D_3$.
\end{example}

Faster implementation of bounds consistency have been designed since Puget's
publication {\cite{Lopez,Melhorn}}. \ Leconte introduced an algorithm that
achieves range consistency {\cite{leconte}}, also based on the Hall's theorem.
\ In dual definitions from above, given a set of variables $K$ let $I_K$ be
the interval $[\min (D_K), \max (D_K)]$ where $D_K$ is the union of all
variable domains; $K$ is a Hall set iff $|K| = |I_K |$.

\begin{proposition}
  The constraint alldifferent alldifferent on variables $x_1, \ldots, x_n$
  where no domain $D_i$ is empty is range consistent iff for each Hall set
  $K$, $D_i \cap I_K = \emptyset$ for all $x_i \nin K$.
\end{proposition}

An algorithm for range consistency can be derived from the previous
proposition in a similar way to the derivation of the bounds consistency
algorithm.

\subsection{Su-Doku as a CSP: a specific problem}

Enumerating Su-Doku grids or solving puzzles within reasonable space and time
limits requires efficient algorithms for a single type of constraint, the
alldifferent constraint. \ While general ``propagate + search''
constraint-solving algorithms will work well on Su-Doku puzzles, specifically
tailored algorithm for the alldifferent constraint work still better. \ There
are several versions of such algorithms, which embody different degrees of
consistency and hence of performance in the puzzle task. \ Modern CSP, SAT and
LP-solvers usually provide specific filtering procedures for the alldifferent
constraint, which make them tools of choice for studying the complexity of the
Su-Doku universe.

\section{Su-Doku as an Exact Cover problem}

In fact, Su-Doku grids and puzzles involve an even more specific constraint
than the alldifferent constraint. \ As is obvious from the previously
mentioned SAT and CSP formulations, the alldifferent constraints in the
Su-Doku problems involve variables having the same domain, namely $M_{n^2}$.
In addition the number of variables in a given constraint is equal to the size
of their shared domain. \ We call such special alldifferent constraint,
{\tmem{permutations}}.

In this section we explore a completely different approach to enumerating and
solving grids and puzzles based on the precedent observation. \ Although the
theoretical basis, going back to matching theory and Hall's results in graph
theory, is the same, the resulting algorithms will significantly differ from
the ones derived from consistency-checking filtering procedures explored in
the previous section of this paper.

\subsection{\label{ExactFormulation}The Exact Cover and Exact Hitting Set
problems}

\begin{definition}
  (Exact Cover) Given a family $\mathfrak{A}=\{A_1, \ldots ., A_m \}$ of
  subsets of a set $X$, $\mathfrak{A}$ is called an {\tmem{exact cover}} of X
  when $\forall x \in X, \exists i, 1 \leqslant i \leqslant m$ such that $x
  \in A_i$
\end{definition}

This definition has also a dual formulation which describes an {\tmem{exact
hitting set}}.

\begin{definition}
  (Distinct Representatives) Given a family $\mathfrak{A}=\{A_1, \ldots ., A_m
  \}$ of subsets of a set $X$, a set of m distinct elements of X, one from
  each $A_i$ is called a {\tmem{set of distinct representatives of
  $\mathfrak{A}$,}} or a hitting set.
  
  If the set of distinct representatives is $X$, it is called an {\tmem{exact
  hitting set}} and $\mathfrak{A}$ is an exact cover of $X$. 
\end{definition}

Finding exact hitting sets and enumerating exact hitting sets may be solved by
backtracking algorithms. \ Their efficiency in this case relies on the fact
that a representative has a unique value among the possible ones in the $A_i$
subset and the filtering procedure, in the previous section sense of CSP
solving, then reduces to the simple elimination of this value from all other
domains.

A {\tmem{permutation problem}} is a constraint satisfaction problem in which
each decision variable takes an unique value, and there is the same number of
values as variables. \ Hence any solution assigns a permutation of the values
to the variables. \ There are $m!$ such permutations for a constraint
involving $m$ variables. \ The important feature of permutation CSPs is that
we can transpose the roles of values and variables in representing the
underlying problem to give a new dual model which is also a permutation
problem. \ Each variable in the original (primal) problem becomes a value in
the dual problem, and vice versa.

An {\tmem{injection problem}} is a CSP in which each decision variable takes a
unique value, and there are more values than variables. (Obviously if there
are fewer values than variables, the problem is trivially unsatisfiable.)

The primal and the dual permutation problems are of course equivalent, but
efficient algorithm can leverage this transposition by switching from one
model to the other when appropriate {\cite{hnich04dual}}. \ Actually the
simple PS-1-2 algorithm described in the previous section indeed used the fact
that the specific alldifferent constraints in Su-Doku problems are all
permutation problems: the so-called \tmtexttt{reduce} functions transpose
values to variables, seeking to further reduce primal and dual domains when
possible.

In Su-Doku problems, each constraint is both a permutation and an exact
hitting set problem as all variables have the same domain.

\paragraph{Matrix representation of exact hitting set and exact cover
problems.}Both exact hitting set and exact cover problem can be represented as
follows. Given a boolean matrix $M$ with $n$ rows and $m$ columns, the problem
is to find a subset $\mathfrak{A}$ of the rows of $M$, such that each column
$j$ in $M$ has exactly one row $i \in \mathfrak{A}$ such that $M_{i, j} = 1$.

For example, consider the following matrix $M$ with $n = 6$ and $m = 4$:

$M = \left(\begin{array}{cccc}
  1 & 0 & 1 & 0\\
  0 & 0 & 1 & 1\\
  0 & 1 & 0 & 0\\
  1 & 0 & 1 & 1\\
  0 & 0 & 0 & 1\\
  1 & 1 & 0 & 0
\end{array}\right)$

Possible exact hitting sets are \{1,3,5\}, \{2,6\} and \{3,4\}.

In the case of permutation problems, the matrices involved are square $n
\times n$ matrices.

\subsection{Enumeration}

The count of exact hitting sets is the number of solutions to the constraints
used in Su-Doku formulations. \ Generally speaking, the number of exact
hitting sets for permutation constraints, i.e. in which the number of values
is the same as variables, is given by the {\tmem{permanent}} of the
representation matrix {\cite{marcus}}.

\begin{definition}
  (Permanent) If $A$ is an n-square matrix then the permanent of $A$ is
  defined by
  \[ \tmop{per} A = \sum_{\sigma \in S_n} \prod_{i = 1}^n a_{i, \sigma (i)} \]
  where the summation extends over $S_n$, the symmetric group of degree $n$.
\end{definition}

The permanent is an appropriate invariant for matrices that arise in
combinatorial investigations where the problem is essentially unaltered by
relabeling of the items under consideration, which is obviously the case in
permutation problems. \ For example, the total number of derangements (``le
probl\`eme des rencontres'') of $n$ distinct items is given by $\tmop{per} (J
- I_n)$ where J is the n-square matrix with every entry equal to 1, and $I_n$
is the n-square identity matrix {\cite{Penrice}}. (This is series A000166 in
the Online Encyclopedia of Integer Sequences.)

Even though the permanent looks superficially likes the more familiar
determinant (without the alternating $\pm$ signs), P\'olya observed that no
uniform affixing of $\pm$ signs to the elements of the matrix can convert the
permanent into the determinant, for $n > 2$. \ The apparent simplification of
definition from the determinant results counter-intuitively in tremendous
complications in the evaluation of permanents. \ In particular, and in
contrast to the determinant, the permanent is not well-behaved under
permutation of rows and columns of the matrix; it is, however, multilinear
like the determinant.

\paragraph{Van der Waerden's Conjecture.}Bounds for the permanent have been
found, however difficult its exact computation turns out to be. Given a
$n$-square matrix $A$, the $i$-th row sum of $A$ is defined by
\[ r_i = \sum_{j = 1}^n a_{i, j} \]
and similarly the $i$-th column sum of $A$ is defined by
\[ c_i = \sum_{j = 1}^n a_{i, j} \]
With these, we introduce a doubly stochastic matrix with this definition.

\begin{definition}
  (Doubly Stochastic Matrix) Let $A = (a_{i, j})$ be a $n$-square matrix, then
  $A$ is doubly stochastic if
  
  $0 \leqslant a_{i, j} \leqslant 1$ for $1 \leqslant i, j \leqslant n$
  
  $r_i = 1$ for $1 \leqslant i \leqslant n$
  
  $c_i = 1$ for $1 \leqslant i \leqslant n$
\end{definition}

Note that the representation matrix of an exact hitting set (or exact cover
problem) is amenable to a doubly stochastic matrix, in the case of
permutation, by replacing each entry equal to 1 with $1 / n$.

Van der Waerden made a conjecture on the lower bound for the permanent of
doubly stochastic matrices in 1926 {\cite{Brualdi}} which was later proved (in
1981) by Egoritchev and by Falikman as exposed by Knuth in
{\cite{KnuthPermanent}}.

\begin{theorem}
  (Van der Waerden's Conjecture) Let $A$ be a doubly stochastic $n$-square
  matrix, then
  \[ \tmop{per} (A) \geqslant \frac{n!}{n^n} \]
  with equality iff $a_{i, j} = 1 / n$ for all $1 \leqslant i, j \leqslant n$.
\end{theorem}

\paragraph{Minc's Conjecture.}There is also a result for the upper bound of
the permanent, due to an conjecture originally due to Minc {\cite{minc63}}. \
The conjecture was first proved by Bregman in 1973 and a simpler proof is due
to Schrijver {\cite{schrijver}}.

\begin{theorem}
  (Minc's Conjecture) Let $A$ be a $n$-square matrix with values in \{0,1\}
  and non-zero sums $r_i$,
  \[ \tmop{per} (A) \leqslant \prod_{i = 1}^n (r_i !)^{1 / r_i} \]
\end{theorem}

There are only few matrices for which an explicit formula for the permanent is
available, the derangements being one instance. \ In fact {\cite{marcus}},
\[ \tmop{per} (z I_n + J) = n! \sum_{r = 0}^n \frac{z^r}{r!} \]
which, for the derangements, gives with $z = - 1$,
\[ \tmop{per} (J - I_n) = n! (1 - \frac{1}{1!} + \frac{1}{2!} - \ldots + (-
   1)^n \frac{1}{n!}) \]
Permanents can be used to evaluate the number of Su-Doku grids. \ The scarce
results known about permanents, however, yield only information on upper
bounds of the number of grids rather than their exact number which was
essentially computed by brute force in {\cite{FJ}}.

\subsection{An implementation of Knuth's ``Dancing Links'' algorithm}

In a famous paper {\cite{knuth00dancing}} Donald Knuth proposed an algorithm
and a very efficient implementation for the exact cover problem. \ While the
paper expands on its application to pentominoes, tetrasticks and to the queens
problems, the algorithm itself, which Knuth called {\tmem{Algorithm
X{\tmem{}}}} ``for lack of a better name'', has a much broader scope. \
Through proper formulation of Su-Doku grid and puzzle problems, it proved
efficient at enumerating grids and solving problems of various sizes.

Knuth's first insight is to point that the matrix representation of the exact
cover or exact hitting set problems makes it a good candidate for
backtracking. \ Algorithm X is a simple expression of a generic backtrack
process.

\paragraph{Knuth's backtracking algorithm for the exact hitting set
problem.}Algorithm X is a nondeterministic algorithm, defined on a given
matrix $A$ of 0s and 1s. \ Citing from Knuth's paper:

\begin{verbatim}
If  is empty, the problem is solved; terminate successfully.
Otherwise choose a column  deterministically.
Choose a row  such that  nondeterministically.
Include  in the partial solution.
For each  such that ,
- delete column  from matrix ;
- for each  such that , delete row  from matrix .
Repeat this algorithm recursively on the reduced matrix .
\end{verbatim}

The nondeterministic choice of a row means that all such rows are successively
(or in parallel) selected for inclusion into the partial solution, the
algorithm proceeding essentially in an independent way on these rows. \ The
choice of the column $c$, on the other hand impacts the execution time and
exploration path of the algorithm. \ Any systematic rule for choosing a column
in the procedure will find all solutions. \ Certain rules, however, work
better than others.

In the Su-Doku experiments we studied two such rules: the {\tmem{random}}
rule, where the column is chosen at random in the reduced matrix, and the
{\tmem{shortest}} rule, where the column having the smaller number of 1s is
selected. \ While for enumeration tasks these options make no real difference,
as Algorithm X in this case behaves basically as a trial and error procedure,
we found that for puzzles, the shortest rule always outperformed the other
one. \ This is also the result of experiments ran on another well-known
combinatorial puzzle, the Langford's problem.

\paragraph{Knuth's ``Dancing Links'' implementation.}In the original paper,
Knuth also proposed a very efficient implementation of Algorithm X based on
doubly-linked circular lists. Each element in the matrix $A$ is represented as
a structured object with pointers to the previous and next elements in the
same row ({\tmem{left}} and {\tmem{right}}), to the previous and next elements
in the same column ({\tmem{up}} and {\tmem{down}}) and an extra-pointer to a
column header structure which keeps track of the column name, its size (the
number of 1s) and additional metric information which can be useful to monitor
the performance of the algorithm.

\begin{verbatim}
typedef struct col {
  struct col *l;
  struct col *r;
  struct cell *u;
  struct cell *d;
  int size;
  char name[8];
  ClientDataPtr clientData;
} *colPtr;

typedef struct cell {
  struct cell *l;
  struct cell *r;
  struct cell *u;
  struct cell *d;
  colPtr c;
} *cellPtr;
\end{verbatim}

The $l$ and $r$ fields of the column headers link remaining columns in the
reduced matrix which need to be covered. \ Global variables point to the
circular list of columns and to the partial solution:

\begin{verbatim}
static struct col S_Header;
static cellPtr *S_Covering;
\end{verbatim}

With these data structures, the concrete implementation of Algorithm X is as
follows:

\begin{verbatim}
search( k ):
If S_Header.r == S_Header, print the current solution and return.
Otherwise choose a column structure .
Cover column .
For each row in  while ,
- set S_Covering[k]=;
- for each  in  while , cover column ;
- search( k+1 );
- set =S_Covering[k], and ;
- for each  in  while , uncover column .
Uncover column  and return.
\end{verbatim}

The search procedure is initially called with k = 0 to enumerate all
solutions.

Knuth's second insight is used to implement the cover/uncover function which
are used to remove and reinstall columns in the matrix. \ Knuth observed that
the ``atomic'' remove operations in a doubly-linked circular list:
\[ x.r.l = x.l \tmop{and} x.l.r = x.r \]
are simply reversed, provided the $x$ data structure is kept intact, by the
subsequent operations:
\[ x.r.l = x \tmop{and} x.l.r = x \]
which will put back $x$ in the circular list.

The cover operation uses the first set of operations to remove a column first
from the header list and then to remove all rows in $c$'s own circular list
from the other column lists they are in:

\begin{verbatim}
int cover( colPtr col ){
  int updates = 0;
  /* Remove col from header list */
  col->r->l = col->l;  col->l->r = col->r;
  updates++;
  
  /* Remove all rows in col list from other col lists they are in */
  cellPtr cell, rowCell;
  for( cell = col->d; cell != (cellPtr)col; cell = cell->d ){
    for( rowCell = cell->r; rowCell != cell; rowCell = rowCell->r ){
      rowCell->d->u = rowCell->u;  rowCell->u->d = rowCell->d;
      rowCell->c->size -= 1;
      updates++;
    }
  }

  return updates;
}
\end{verbatim}

The function also keeps track of counters for statistical purposes and
decrements the column size in the header. \ The uncovering operation is
symmetric, taking place in precisely the reverse order of the covering
operation:

\begin{verbatim}
/*
 * uncover - Inverse cover
 */
void uncover( colPtr col ){
  /* Inserts all row cells */
  cellPtr cell, rowCell;
  for( cell = col->u; cell != (cellPtr)col; cell = cell->u ){
    for( rowCell = cell->l; rowCell != cell; rowCell= rowCell->l ){
      rowCell->c->size += 1;
      rowCell->d->u = rowCell;  rowCell->u->d = rowCell;
    }
  }

  /* Inserts in header list */
  col->r->l = col;  col->l->r = col;
}
\end{verbatim}

The disconnected then reconnected links perform what Knuth called a ``dance''
which gave its name to this implementation known as the ``Dancing Links''.

The running time of the algorithm is essentially proportional to the number of
times it applies the remove operation, counted here with the
\tmtexttt{updates\tmtexttt{}} variable. \ It is possible to get good estimates
of the running time on average by running the above procedure a few times and
applying techniques described elsewhere by Knuth {\cite{knuth75EEB}} and
Hammersley and Morton {\cite{hammersely54}} (so called ``Poor Man's Monte
Carlo'').

\paragraph{Deriving cover matrices from representation matrices.}We now turn
to the proper formulation of Su-Doku questions for calculations by the Dancing
Links algorithm. \ Considering an elementary constraint in the size 2 Su-Doku
puzzle, for instance on one row of the 4x4 grid, its matrix representation as
in \ref{ExactFormulation} is simply:
\[ M = \left(\begin{array}{cccc}
     1 & 1 & 1 & 1\\
     1 & 1 & 1 & 1\\
     1 & 1 & 1 & 1\\
     1 & 1 & 1 & 1
   \end{array}\right) \]
In $M$ each column stands for a variable in the alldifferent constraint; there
are four of them, one for each cell in the grid's row under consideration. \
The rows represent each of the possible four values $1, 2 \ldots 4$ of all
these variables. \ Remember that in a permutation constraint all variable
domains are the same and their size is equal to the number of variables, here
four.

And of course, should we be interested in a single permutation constraint, the
number of solutions, which as mentioned above is expressed by the permanent of
this ``all-1s'' matrix, is simply, in this case, the number of permutations of
four elements, i.e. $4! = 24$.

Now in order to obtain the $A$ matrix for the Dancing Links algorithm, we
augment the matrix $M$ with the fact that each variable must have only one
value. \ This is captured by four additional columns, one for each variable,
containing a 1 for a given (row) value assigned to the variable and 0
otherwise:
\[ \begin{array}{lllllllll}
     & x_1 & x_2 & x_3 & x_4 & C_1 & C_2 & C_3 & C_4\\
     x = 1 & 1 & 0 & 0 & 0 & 1 & 0 & 0 & 0\\
     x_{} = 1 & 1 & 0 & 0 & 0 & 0 & 1 & 0 & 0\\
     x_{} = 1 & 1 & 0 & 0 & 0 & 0 & 0 & 1 & 0\\
     x = 1 & 1 & 0 & 0 & 0 & 0 & 0 & 0 & 1\\
     x = 2 & 0 & 1 & 0 & 0 & 1 & 0 & 0 & 0\\
     x = 2 & 0 & 1 & 0 & 0 & 0 & 1 & 0 & 0\\
     x = 2 & 0 & 1 & 0 & 0 & 0 & 0 & 1 & 0\\
     x = 2 & 0 & 1 & 0 & 0 & 0 & 0 & 0 & 1\\
     x = 3 & 0 & 0 & 1 & 0 & 1 & 0 & 0 & 0\\
     x = 3 & 0 & 0 & 1 & 0 & 0 & 1 & 0 & 0\\
     x = 3 & 0 & 0 & 1 & 0 & 0 & 0 & 1 & 0\\
     x = 3 & 0 & 0 & 1 & 0 & 0 & 0 & 0 & 1\\
     x = 4 & 0 & 0 & 0 & 1 & 1 & 0 & 0 & 0\\
     x = 4 & 0 & 0 & 0 & 1 & 0 & 1 & 0 & 0\\
     x = 4 & 0 & 0 & 0 & 1 & 0 & 0 & 1 & 0\\
     x = 4 & 0 & 0 & 0 & 1 & 0 & 0 & 0 & 1
   \end{array} \]
In this table the $x_i \tmop{column}$ is the $i$-th variable in the $C$
permutation constraint, the $C_i$ column represents the $i$-th position in the
constraint and each row a value from the shared domain $\{1, 2, 3, 4\}$. \ The
$A$ matrix has 8 columns and 16 rows.

More generally speaking, for a size $n$ permutation constraint the $A$ matrix
counts $n^2$ rows and $n + n = 2 n$ columns.

For a complete Su-Doku grid, there is one such constraint per line, per file
and per block. \ In addition, variables are shared between constaints, each
one appearing in 3 constraints. \ Let us consider a Su-Doku of size $n$, which
contains $n^4$, cells in $n^2$ lines by $n^2$ files, and $n^2$ blocks. \ The
full size $A$ matrix for the Dancing Links algorithm has $n^4 + n^4 + n^4 +
n^4 = 4 n^4$ columns, one for each of the cells, and $n^2$ for each of the
line, file and block in the grid. It also has $n^6$ rows, one row for each of
the $n^2$ possible value for each of the $n^4$ cells. \ The following table
indicate the matrices sizes for different Su-Doku problems:
\[ \begin{array}{ll}
     n & \tmop{Matrix} (\tmop{rows} \times \tmop{cols})\\
     2 & 64 \times 64\\
     3 & 729 \times 324\\
     4 & 4096 \times 1024
   \end{array} \]

These sizes are small enough for the algorithm to perform satisfactorily on
modern PCs.

\paragraph{Enumerating grids and solving puzzles with the ``Dancing
Links''.}Having augmented the matrix to prepare it for the Dancing Links
algorithms we are now ready to put the algorithm through different chores.

In order to enumerate all Su-Doku valid grids we simply run the
\tmtexttt{search( 0 )} procedure with the appropriate $A$ matrix as above. \
Of course, while the 288 solutions of the size 2 Su-Doku grids are quickly
enumerated, the size 3 grid takes evidently too long to list. \ Interestingly
enough, size 2 variations of Su-Doku grids, such as {\tmem{diagonal Su-Doku
grids}} where in addition one requires that all numbers in both diagonals to
be different -- adding two additional permutation constraints to the existing
set, captured by $2 n^2$ additional columns in the $A$ matrix -- can also be
enumerated by the same procedure.

In order to solve puzzles, we need to remove from matrix $A$ the rows
corresponding to the givens in the puzzle. \ In our implementation, this is
simply another parameter file to the command line. \ If there are $k$ such
givens in the puzzle, $k$ rows are initially added to the partial solution and
the procedure \tmtexttt{search( k )} is called. \ The algorithm then proceeds,
as above, to enumerate all solutions to the puzzle. \ It can be used to
validate a puzzle, making sure that it has only one solution.

\subsection{Experimentation and results}

\paragraph{Enumerating size-2 Su-Doku grids.}Running the Dancing Links
algorithm on the 64 by 64 size-2 Su-Doku $A$ matrix, produces the first of the
288 solutions almost immediately:

\tmtexttt{Read 64 columns from sud2.mat

Read 64 rows from file sud2.mat

[16] New covering 1/1 in 0 secs, 0 usecs:

\ \ \ \ Depth \ \ \ \ \ \ \ \ \ Covers \ \ \ \ \ Backtracks \ \ \ \ \ \ \ \
Degrees

\ \ \ \ \ \ \ 0 \ \ \ \ \ \ \ \ \ \ \ \ \ 37 \ \ \ \ \ \ \ \ \ \ \ \ \ \ 1 \
\ \ \ \ \ \ \ \ \ \ \ \ \ 4

\ \ \ \ \ \ \ 1 \ \ \ \ \ \ \ \ \ \ \ \ \ 25 \ \ \ \ \ \ \ \ \ \ \ \ \ \ 1 \
\ \ \ \ \ \ \ \ \ \ \ \ \ 2

\ \ \ \ \ \ \ 2 \ \ \ \ \ \ \ \ \ \ \ \ \ 22 \ \ \ \ \ \ \ \ \ \ \ \ \ \ 1 \
\ \ \ \ \ \ \ \ \ \ \ \ \ 2

\ \ \ \ \ \ \ 3 \ \ \ \ \ \ \ \ \ \ \ \ \ 16 \ \ \ \ \ \ \ \ \ \ \ \ \ \ 1 \
\ \ \ \ \ \ \ \ \ \ \ \ \ 1

\ \ \ \ \ \ \ 4 \ \ \ \ \ \ \ \ \ \ \ \ \ 28 \ \ \ \ \ \ \ \ \ \ \ \ \ \ 1 \
\ \ \ \ \ \ \ \ \ \ \ \ \ 3

\ \ \ \ \ \ \ 5 \ \ \ \ \ \ \ \ \ \ \ \ \ 16 \ \ \ \ \ \ \ \ \ \ \ \ \ \ 1 \
\ \ \ \ \ \ \ \ \ \ \ \ \ 1

\ \ \ \ \ \ \ 6 \ \ \ \ \ \ \ \ \ \ \ \ \ 19 \ \ \ \ \ \ \ \ \ \ \ \ \ \ 1 \
\ \ \ \ \ \ \ \ \ \ \ \ \ 2

\ \ \ \ \ \ \ 7 \ \ \ \ \ \ \ \ \ \ \ \ \ 10 \ \ \ \ \ \ \ \ \ \ \ \ \ \ 1 \
\ \ \ \ \ \ \ \ \ \ \ \ \ 1

\ \ \ \ \ \ \ 8 \ \ \ \ \ \ \ \ \ \ \ \ \ 10 \ \ \ \ \ \ \ \ \ \ \ \ \ \ 1 \
\ \ \ \ \ \ \ \ \ \ \ \ \ 1

\ \ \ \ \ \ \ 9 \ \ \ \ \ \ \ \ \ \ \ \ \ 16 \ \ \ \ \ \ \ \ \ \ \ \ \ \ 1 \
\ \ \ \ \ \ \ \ \ \ \ \ \ 2

\ \ \ \ \ \ 10 \ \ \ \ \ \ \ \ \ \ \ \ \ 10 \ \ \ \ \ \ \ \ \ \ \ \ \ \ 1 \ \
\ \ \ \ \ \ \ \ \ \ \ \ 1

\ \ \ \ \ \ 11 \ \ \ \ \ \ \ \ \ \ \ \ \ \ 7 \ \ \ \ \ \ \ \ \ \ \ \ \ \ 1 \
\ \ \ \ \ \ \ \ \ \ \ \ \ 1

\ \ \ \ \ \ 12 \ \ \ \ \ \ \ \ \ \ \ \ \ 16 \ \ \ \ \ \ \ \ \ \ \ \ \ \ 1 \ \
\ \ \ \ \ \ \ \ \ \ \ \ 2

\ \ \ \ \ \ 13 \ \ \ \ \ \ \ \ \ \ \ \ \ 10 \ \ \ \ \ \ \ \ \ \ \ \ \ \ 1 \ \
\ \ \ \ \ \ \ \ \ \ \ \ 1

\ \ \ \ \ \ 14 \ \ \ \ \ \ \ \ \ \ \ \ \ \ 7 \ \ \ \ \ \ \ \ \ \ \ \ \ \ 1 \
\ \ \ \ \ \ \ \ \ \ \ \ \ 1

\ \ \ \ \ \ 15 \ \ \ \ \ \ \ \ \ \ \ \ \ \ 7 \ \ \ \ \ \ \ \ \ \ \ \ \ \ 1 \
\ \ \ \ \ \ \ \ \ \ \ \ \ 1

\ \ \ \ Total \ \ \ \ \ \ \ \ \ \ \ 256 \ \ \ \ \ \ \ \ \ \ \ \ \ 16

Estimation of solution path:

7620

}

The \tmtexttt{sud2.mat} file is the $A$ matrix for the size-2 Su-Doku grid. \
The trace table shows the depth, i.e. the value of $k$ which indicates the
depth in the backtrack tree; the cover count, which is the number of
elementary remove operations in the circular lists; the number of backtracking
steps at each depth level; and the degree, the number of children nodes
explored at each level. \ Finally the estimation of the average number of
operations to reach a solution is printed according to the ``Poor Man's Monte
Carlo'' method.

\paragraph{Counting Su-Doku grids.}The algorithm can be used to count the
number of Su-Doku grids, here for the size-2 grid:

\tmtexttt{Read 64 columns from sud2.mat

Read 64 rows from file sud2.mat

1 \ \ \ \ \ \ 16 \ \ \ \ \ \ \ \ \ \ \ 7620 \ \ \ \ \ \ \ \ \ \ \ 7620

2 \ \ \ \ \ \ 16 \ \ \ \ \ \ \ \ \ \ \ 7620 \ \ \ \ \ \ \ \ \ \ 15240

3 \ \ \ \ \ \ 16 \ \ \ \ \ \ \ \ \ \ \ 5316 \ \ \ \ \ \ \ \ \ \ 20556

4 \ \ \ \ \ \ 16 \ \ \ \ \ \ \ \ \ \ \ 5316 \ \ \ \ \ \ \ \ \ \ 25872

5 \ \ \ \ \ \ 16 \ \ \ \ \ \ \ \ \ \ \ 7620 \ \ \ \ \ \ \ \ \ \ 33492

6 \ \ \ \ \ \ 16 \ \ \ \ \ \ \ \ \ \ \ 7620 \ \ \ \ \ \ \ \ \ \ 41112

7 \ \ \ \ \ \ 16 \ \ \ \ \ \ \ \ \ \ \ 7620 \ \ \ \ \ \ \ \ \ \ 48732

8 \ \ \ \ \ \ 16 \ \ \ \ \ \ \ \ \ \ \ 7620 \ \ \ \ \ \ \ \ \ \ 56352

9 \ \ \ \ \ \ 16 \ \ \ \ \ \ \ \ \ \ \ 5316 \ \ \ \ \ \ \ \ \ \ 61668

10 \ \ \ \ \ 16 \ \ \ \ \ \ \ \ \ \ \ 5316 \ \ \ \ \ \ \ \ \ \ 66984

11 \ \ \ \ \ 16 \ \ \ \ \ \ \ \ \ \ \ 7620 \ \ \ \ \ \ \ \ \ \ 74604

12 \ \ \ \ \ 16 \ \ \ \ \ \ \ \ \ \ \ 7620 \ \ \ \ \ \ \ \ \ \ 82224

13 \ \ \ \ \ 16 \ \ \ \ \ \ \ \ \ \ \ 7620 \ \ \ \ \ \ \ \ \ \ 89844

14 \ \ \ \ \ 16 \ \ \ \ \ \ \ \ \ \ \ 7620 \ \ \ \ \ \ \ \ \ \ 97464

15 \ \ \ \ \ 16 \ \ \ \ \ \ \ \ \ \ \ 5316 \ \ \ \ \ \ \ \ \ 102780

16 \ \ \ \ \ 16 \ \ \ \ \ \ \ \ \ \ \ 5316 \ \ \ \ \ \ \ \ \ 108096

17 \ \ \ \ \ 16 \ \ \ \ \ \ \ \ \ \ \ 7620 \ \ \ \ \ \ \ \ \ 115716

18 \ \ \ \ \ 16 \ \ \ \ \ \ \ \ \ \ \ 7620 \ \ \ \ \ \ \ \ \ 123336

19 \ \ \ \ \ 16 \ \ \ \ \ \ \ \ \ \ \ 7620 \ \ \ \ \ \ \ \ \ 130956

20 \ \ \ \ \ 16 \ \ \ \ \ \ \ \ \ \ \ 7620 \ \ \ \ \ \ \ \ \ 138576

21 \ \ \ \ \ 16 \ \ \ \ \ \ \ \ \ \ \ 5316 \ \ \ \ \ \ \ \ \ 143892

22 \ \ \ \ \ 16 \ \ \ \ \ \ \ \ \ \ \ 5316 \ \ \ \ \ \ \ \ \ 149208

23 \ \ \ \ \ 16 \ \ \ \ \ \ \ \ \ \ \ 7620 \ \ \ \ \ \ \ \ \ 156828

24 \ \ \ \ \ 16 \ \ \ \ \ \ \ \ \ \ \ 7620 \ \ \ \ \ \ \ \ \ 164448

25 \ \ \ \ \ 16 \ \ \ \ \ \ \ \ \ \ \ 7620 \ \ \ \ \ \ \ \ \ 172068

26 \ \ \ \ \ 16 \ \ \ \ \ \ \ \ \ \ \ 7620 \ \ \ \ \ \ \ \ \ 179688

27 \ \ \ \ \ 16 \ \ \ \ \ \ \ \ \ \ \ 5316 \ \ \ \ \ \ \ \ \ 185004

28 \ \ \ \ \ 16 \ \ \ \ \ \ \ \ \ \ \ 5316 \ \ \ \ \ \ \ \ \ 190320

29 \ \ \ \ \ 16 \ \ \ \ \ \ \ \ \ \ \ 7620 \ \ \ \ \ \ \ \ \ 197940

30 \ \ \ \ \ 16 \ \ \ \ \ \ \ \ \ \ \ 7620 \ \ \ \ \ \ \ \ \ 205560

31 \ \ \ \ \ 16 \ \ \ \ \ \ \ \ \ \ \ 7620 \ \ \ \ \ \ \ \ \ 213180

32 \ \ \ \ \ 16 \ \ \ \ \ \ \ \ \ \ \ 7620 \ \ \ \ \ \ \ \ \ 220800

33 \ \ \ \ \ 16 \ \ \ \ \ \ \ \ \ \ \ 5316 \ \ \ \ \ \ \ \ \ 226116

34 \ \ \ \ \ 16 \ \ \ \ \ \ \ \ \ \ \ 5316 \ \ \ \ \ \ \ \ \ 231432

35 \ \ \ \ \ 16 \ \ \ \ \ \ \ \ \ \ \ 7620 \ \ \ \ \ \ \ \ \ 239052

36 \ \ \ \ \ 16 \ \ \ \ \ \ \ \ \ \ \ 7620 \ \ \ \ \ \ \ \ \ 246672

37 \ \ \ \ \ 16 \ \ \ \ \ \ \ \ \ \ \ 5316 \ \ \ \ \ \ \ \ \ 251988

38 \ \ \ \ \ 16 \ \ \ \ \ \ \ \ \ \ \ 7620 \ \ \ \ \ \ \ \ \ 259608

39 \ \ \ \ \ 16 \ \ \ \ \ \ \ \ \ \ \ 7620 \ \ \ \ \ \ \ \ \ 267228

40 \ \ \ \ \ 16 \ \ \ \ \ \ \ \ \ \ \ 7620 \ \ \ \ \ \ \ \ \ 274848

41 \ \ \ \ \ 16 \ \ \ \ \ \ \ \ \ \ \ 7620 \ \ \ \ \ \ \ \ \ 282468

42 \ \ \ \ \ 16 \ \ \ \ \ \ \ \ \ \ \ 5316 \ \ \ \ \ \ \ \ \ 287784

43 \ \ \ \ \ 16 \ \ \ \ \ \ \ \ \ \ \ 5316 \ \ \ \ \ \ \ \ \ 293100

44 \ \ \ \ \ 16 \ \ \ \ \ \ \ \ \ \ \ 7620 \ \ \ \ \ \ \ \ \ 300720

45 \ \ \ \ \ 16 \ \ \ \ \ \ \ \ \ \ \ 7620 \ \ \ \ \ \ \ \ \ 308340

46 \ \ \ \ \ 16 \ \ \ \ \ \ \ \ \ \ \ 7620 \ \ \ \ \ \ \ \ \ 315960

47 \ \ \ \ \ 16 \ \ \ \ \ \ \ \ \ \ \ 7620 \ \ \ \ \ \ \ \ \ 323580

48 \ \ \ \ \ 16 \ \ \ \ \ \ \ \ \ \ \ 5316 \ \ \ \ \ \ \ \ \ 328896

49 \ \ \ \ \ 16 \ \ \ \ \ \ \ \ \ \ \ 5316 \ \ \ \ \ \ \ \ \ 334212

50 \ \ \ \ \ 16 \ \ \ \ \ \ \ \ \ \ \ 7620 \ \ \ \ \ \ \ \ \ 341832

51 \ \ \ \ \ 16 \ \ \ \ \ \ \ \ \ \ \ 7620 \ \ \ \ \ \ \ \ \ 349452

52 \ \ \ \ \ 16 \ \ \ \ \ \ \ \ \ \ \ 7620 \ \ \ \ \ \ \ \ \ 357072

53 \ \ \ \ \ 16 \ \ \ \ \ \ \ \ \ \ \ 7620 \ \ \ \ \ \ \ \ \ 364692

54 \ \ \ \ \ 16 \ \ \ \ \ \ \ \ \ \ \ 5316 \ \ \ \ \ \ \ \ \ 370008

55 \ \ \ \ \ 16 \ \ \ \ \ \ \ \ \ \ \ 5316 \ \ \ \ \ \ \ \ \ 375324

56 \ \ \ \ \ 16 \ \ \ \ \ \ \ \ \ \ \ 7620 \ \ \ \ \ \ \ \ \ 382944

57 \ \ \ \ \ 16 \ \ \ \ \ \ \ \ \ \ \ 7620 \ \ \ \ \ \ \ \ \ 390564

58 \ \ \ \ \ 16 \ \ \ \ \ \ \ \ \ \ \ 7620 \ \ \ \ \ \ \ \ \ 398184

59 \ \ \ \ \ 16 \ \ \ \ \ \ \ \ \ \ \ 7620 \ \ \ \ \ \ \ \ \ 405804

60 \ \ \ \ \ 16 \ \ \ \ \ \ \ \ \ \ \ 5316 \ \ \ \ \ \ \ \ \ 411120

61 \ \ \ \ \ 16 \ \ \ \ \ \ \ \ \ \ \ 5316 \ \ \ \ \ \ \ \ \ 416436

62 \ \ \ \ \ 16 \ \ \ \ \ \ \ \ \ \ \ 7620 \ \ \ \ \ \ \ \ \ 424056

63 \ \ \ \ \ 16 \ \ \ \ \ \ \ \ \ \ \ 7620 \ \ \ \ \ \ \ \ \ 431676

64 \ \ \ \ \ 16 \ \ \ \ \ \ \ \ \ \ \ 7620 \ \ \ \ \ \ \ \ \ 439296

65 \ \ \ \ \ 16 \ \ \ \ \ \ \ \ \ \ \ 7620 \ \ \ \ \ \ \ \ \ 446916

66 \ \ \ \ \ 16 \ \ \ \ \ \ \ \ \ \ \ 5316 \ \ \ \ \ \ \ \ \ 452232

67 \ \ \ \ \ 16 \ \ \ \ \ \ \ \ \ \ \ 5316 \ \ \ \ \ \ \ \ \ 457548

68 \ \ \ \ \ 16 \ \ \ \ \ \ \ \ \ \ \ 7620 \ \ \ \ \ \ \ \ \ 465168

69 \ \ \ \ \ 16 \ \ \ \ \ \ \ \ \ \ \ 7620 \ \ \ \ \ \ \ \ \ 472788

70 \ \ \ \ \ 16 \ \ \ \ \ \ \ \ \ \ \ 7620 \ \ \ \ \ \ \ \ \ 480408

71 \ \ \ \ \ 16 \ \ \ \ \ \ \ \ \ \ \ 7620 \ \ \ \ \ \ \ \ \ 488028

72 \ \ \ \ \ 16 \ \ \ \ \ \ \ \ \ \ \ 5316 \ \ \ \ \ \ \ \ \ 493344

73 \ \ \ \ \ 16 \ \ \ \ \ \ \ \ \ \ \ 7620 \ \ \ \ \ \ \ \ \ 500964

74 \ \ \ \ \ 16 \ \ \ \ \ \ \ \ \ \ \ 7620 \ \ \ \ \ \ \ \ \ 508584

75 \ \ \ \ \ 16 \ \ \ \ \ \ \ \ \ \ \ 5316 \ \ \ \ \ \ \ \ \ 513900

76 \ \ \ \ \ 16 \ \ \ \ \ \ \ \ \ \ \ 5316 \ \ \ \ \ \ \ \ \ 519216

77 \ \ \ \ \ 16 \ \ \ \ \ \ \ \ \ \ \ 7620 \ \ \ \ \ \ \ \ \ 526836

78 \ \ \ \ \ 16 \ \ \ \ \ \ \ \ \ \ \ 7620 \ \ \ \ \ \ \ \ \ 534456

79 \ \ \ \ \ 16 \ \ \ \ \ \ \ \ \ \ \ 7620 \ \ \ \ \ \ \ \ \ 542076

80 \ \ \ \ \ 16 \ \ \ \ \ \ \ \ \ \ \ 7620 \ \ \ \ \ \ \ \ \ 549696

81 \ \ \ \ \ 16 \ \ \ \ \ \ \ \ \ \ \ 5316 \ \ \ \ \ \ \ \ \ 555012

82 \ \ \ \ \ 16 \ \ \ \ \ \ \ \ \ \ \ 5316 \ \ \ \ \ \ \ \ \ 560328

83 \ \ \ \ \ 16 \ \ \ \ \ \ \ \ \ \ \ 7620 \ \ \ \ \ \ \ \ \ 567948

84 \ \ \ \ \ 16 \ \ \ \ \ \ \ \ \ \ \ 7620 \ \ \ \ \ \ \ \ \ 575568

85 \ \ \ \ \ 16 \ \ \ \ \ \ \ \ \ \ \ 7620 \ \ \ \ \ \ \ \ \ 583188

86 \ \ \ \ \ 16 \ \ \ \ \ \ \ \ \ \ \ 7620 \ \ \ \ \ \ \ \ \ 590808

87 \ \ \ \ \ 16 \ \ \ \ \ \ \ \ \ \ \ 5316 \ \ \ \ \ \ \ \ \ 596124

88 \ \ \ \ \ 16 \ \ \ \ \ \ \ \ \ \ \ 5316 \ \ \ \ \ \ \ \ \ 601440

89 \ \ \ \ \ 16 \ \ \ \ \ \ \ \ \ \ \ 7620 \ \ \ \ \ \ \ \ \ 609060

90 \ \ \ \ \ 16 \ \ \ \ \ \ \ \ \ \ \ 7620 \ \ \ \ \ \ \ \ \ 616680

91 \ \ \ \ \ 16 \ \ \ \ \ \ \ \ \ \ \ 7620 \ \ \ \ \ \ \ \ \ 624300

92 \ \ \ \ \ 16 \ \ \ \ \ \ \ \ \ \ \ 7620 \ \ \ \ \ \ \ \ \ 631920

93 \ \ \ \ \ 16 \ \ \ \ \ \ \ \ \ \ \ 5316 \ \ \ \ \ \ \ \ \ 637236

94 \ \ \ \ \ 16 \ \ \ \ \ \ \ \ \ \ \ 5316 \ \ \ \ \ \ \ \ \ 642552

95 \ \ \ \ \ 16 \ \ \ \ \ \ \ \ \ \ \ 7620 \ \ \ \ \ \ \ \ \ 650172

96 \ \ \ \ \ 16 \ \ \ \ \ \ \ \ \ \ \ 7620 \ \ \ \ \ \ \ \ \ 657792

97 \ \ \ \ \ 16 \ \ \ \ \ \ \ \ \ \ \ 7620 \ \ \ \ \ \ \ \ \ 665412

98 \ \ \ \ \ 16 \ \ \ \ \ \ \ \ \ \ \ 7620 \ \ \ \ \ \ \ \ \ 673032

99 \ \ \ \ \ 16 \ \ \ \ \ \ \ \ \ \ \ 5316 \ \ \ \ \ \ \ \ \ 678348

100 \ \ \ \ 16 \ \ \ \ \ \ \ \ \ \ \ 5316 \ \ \ \ \ \ \ \ \ 683664

101 \ \ \ \ 16 \ \ \ \ \ \ \ \ \ \ \ 7620 \ \ \ \ \ \ \ \ \ 691284

102 \ \ \ \ 16 \ \ \ \ \ \ \ \ \ \ \ 7620 \ \ \ \ \ \ \ \ \ 698904

103 \ \ \ \ 16 \ \ \ \ \ \ \ \ \ \ \ 7620 \ \ \ \ \ \ \ \ \ 706524

104 \ \ \ \ 16 \ \ \ \ \ \ \ \ \ \ \ 7620 \ \ \ \ \ \ \ \ \ 714144

105 \ \ \ \ 16 \ \ \ \ \ \ \ \ \ \ \ 5316 \ \ \ \ \ \ \ \ \ 719460

106 \ \ \ \ 16 \ \ \ \ \ \ \ \ \ \ \ 5316 \ \ \ \ \ \ \ \ \ 724776

107 \ \ \ \ 16 \ \ \ \ \ \ \ \ \ \ \ 7620 \ \ \ \ \ \ \ \ \ 732396

108 \ \ \ \ 16 \ \ \ \ \ \ \ \ \ \ \ 7620 \ \ \ \ \ \ \ \ \ 740016

109 \ \ \ \ 16 \ \ \ \ \ \ \ \ \ \ \ 5316 \ \ \ \ \ \ \ \ \ 745332

110 \ \ \ \ 16 \ \ \ \ \ \ \ \ \ \ \ 7620 \ \ \ \ \ \ \ \ \ 752952

111 \ \ \ \ 16 \ \ \ \ \ \ \ \ \ \ \ 7620 \ \ \ \ \ \ \ \ \ 760572

112 \ \ \ \ 16 \ \ \ \ \ \ \ \ \ \ \ 7620 \ \ \ \ \ \ \ \ \ 768192

113 \ \ \ \ 16 \ \ \ \ \ \ \ \ \ \ \ 7620 \ \ \ \ \ \ \ \ \ 775812

114 \ \ \ \ 16 \ \ \ \ \ \ \ \ \ \ \ 5316 \ \ \ \ \ \ \ \ \ 781128

115 \ \ \ \ 16 \ \ \ \ \ \ \ \ \ \ \ 5316 \ \ \ \ \ \ \ \ \ 786444

116 \ \ \ \ 16 \ \ \ \ \ \ \ \ \ \ \ 7620 \ \ \ \ \ \ \ \ \ 794064

117 \ \ \ \ 16 \ \ \ \ \ \ \ \ \ \ \ 7620 \ \ \ \ \ \ \ \ \ 801684

118 \ \ \ \ 16 \ \ \ \ \ \ \ \ \ \ \ 7620 \ \ \ \ \ \ \ \ \ 809304

119 \ \ \ \ 16 \ \ \ \ \ \ \ \ \ \ \ 7620 \ \ \ \ \ \ \ \ \ 816924

120 \ \ \ \ 16 \ \ \ \ \ \ \ \ \ \ \ 5316 \ \ \ \ \ \ \ \ \ 822240

121 \ \ \ \ 16 \ \ \ \ \ \ \ \ \ \ \ 5316 \ \ \ \ \ \ \ \ \ 827556

122 \ \ \ \ 16 \ \ \ \ \ \ \ \ \ \ \ 7620 \ \ \ \ \ \ \ \ \ 835176

123 \ \ \ \ 16 \ \ \ \ \ \ \ \ \ \ \ 7620 \ \ \ \ \ \ \ \ \ 842796

124 \ \ \ \ 16 \ \ \ \ \ \ \ \ \ \ \ 7620 \ \ \ \ \ \ \ \ \ 850416

125 \ \ \ \ 16 \ \ \ \ \ \ \ \ \ \ \ 7620 \ \ \ \ \ \ \ \ \ 858036

126 \ \ \ \ 16 \ \ \ \ \ \ \ \ \ \ \ 5316 \ \ \ \ \ \ \ \ \ 863352

127 \ \ \ \ 16 \ \ \ \ \ \ \ \ \ \ \ 5316 \ \ \ \ \ \ \ \ \ 868668

128 \ \ \ \ 16 \ \ \ \ \ \ \ \ \ \ \ 7620 \ \ \ \ \ \ \ \ \ 876288

129 \ \ \ \ 16 \ \ \ \ \ \ \ \ \ \ \ 7620 \ \ \ \ \ \ \ \ \ 883908

130 \ \ \ \ 16 \ \ \ \ \ \ \ \ \ \ \ 7620 \ \ \ \ \ \ \ \ \ 891528

131 \ \ \ \ 16 \ \ \ \ \ \ \ \ \ \ \ 7620 \ \ \ \ \ \ \ \ \ 899148

132 \ \ \ \ 16 \ \ \ \ \ \ \ \ \ \ \ 5316 \ \ \ \ \ \ \ \ \ 904464

133 \ \ \ \ 16 \ \ \ \ \ \ \ \ \ \ \ 5316 \ \ \ \ \ \ \ \ \ 909780

134 \ \ \ \ 16 \ \ \ \ \ \ \ \ \ \ \ 7620 \ \ \ \ \ \ \ \ \ 917400

135 \ \ \ \ 16 \ \ \ \ \ \ \ \ \ \ \ 7620 \ \ \ \ \ \ \ \ \ 925020

136 \ \ \ \ 16 \ \ \ \ \ \ \ \ \ \ \ 7620 \ \ \ \ \ \ \ \ \ 932640

137 \ \ \ \ 16 \ \ \ \ \ \ \ \ \ \ \ 7620 \ \ \ \ \ \ \ \ \ 940260

138 \ \ \ \ 16 \ \ \ \ \ \ \ \ \ \ \ 5316 \ \ \ \ \ \ \ \ \ 945576

139 \ \ \ \ 16 \ \ \ \ \ \ \ \ \ \ \ 5316 \ \ \ \ \ \ \ \ \ 950892

140 \ \ \ \ 16 \ \ \ \ \ \ \ \ \ \ \ 7620 \ \ \ \ \ \ \ \ \ 958512

141 \ \ \ \ 16 \ \ \ \ \ \ \ \ \ \ \ 7620 \ \ \ \ \ \ \ \ \ 966132

142 \ \ \ \ 16 \ \ \ \ \ \ \ \ \ \ \ 7620 \ \ \ \ \ \ \ \ \ 973752

143 \ \ \ \ 16 \ \ \ \ \ \ \ \ \ \ \ 7620 \ \ \ \ \ \ \ \ \ 981372

144 \ \ \ \ 16 \ \ \ \ \ \ \ \ \ \ \ 5316 \ \ \ \ \ \ \ \ \ 986688

145 \ \ \ \ 16 \ \ \ \ \ \ \ \ \ \ \ 7620 \ \ \ \ \ \ \ \ \ 994308

146 \ \ \ \ 16 \ \ \ \ \ \ \ \ \ \ \ 7620 \ \ \ \ \ \ \ \ 1001928

147 \ \ \ \ 16 \ \ \ \ \ \ \ \ \ \ \ 5316 \ \ \ \ \ \ \ \ 1007244

148 \ \ \ \ 16 \ \ \ \ \ \ \ \ \ \ \ 5316 \ \ \ \ \ \ \ \ 1012560

149 \ \ \ \ 16 \ \ \ \ \ \ \ \ \ \ \ 7620 \ \ \ \ \ \ \ \ 1020180

150 \ \ \ \ 16 \ \ \ \ \ \ \ \ \ \ \ 7620 \ \ \ \ \ \ \ \ 1027800

151 \ \ \ \ 16 \ \ \ \ \ \ \ \ \ \ \ 7620 \ \ \ \ \ \ \ \ 1035420

152 \ \ \ \ 16 \ \ \ \ \ \ \ \ \ \ \ 7620 \ \ \ \ \ \ \ \ 1043040

153 \ \ \ \ 16 \ \ \ \ \ \ \ \ \ \ \ 5316 \ \ \ \ \ \ \ \ 1048356

154 \ \ \ \ 16 \ \ \ \ \ \ \ \ \ \ \ 5316 \ \ \ \ \ \ \ \ 1053672

155 \ \ \ \ 16 \ \ \ \ \ \ \ \ \ \ \ 7620 \ \ \ \ \ \ \ \ 1061292

156 \ \ \ \ 16 \ \ \ \ \ \ \ \ \ \ \ 7620 \ \ \ \ \ \ \ \ 1068912

157 \ \ \ \ 16 \ \ \ \ \ \ \ \ \ \ \ 7620 \ \ \ \ \ \ \ \ 1076532

158 \ \ \ \ 16 \ \ \ \ \ \ \ \ \ \ \ 7620 \ \ \ \ \ \ \ \ 1084152

159 \ \ \ \ 16 \ \ \ \ \ \ \ \ \ \ \ 5316 \ \ \ \ \ \ \ \ 1089468

160 \ \ \ \ 16 \ \ \ \ \ \ \ \ \ \ \ 5316 \ \ \ \ \ \ \ \ 1094784

161 \ \ \ \ 16 \ \ \ \ \ \ \ \ \ \ \ 7620 \ \ \ \ \ \ \ \ 1102404

162 \ \ \ \ 16 \ \ \ \ \ \ \ \ \ \ \ 7620 \ \ \ \ \ \ \ \ 1110024

163 \ \ \ \ 16 \ \ \ \ \ \ \ \ \ \ \ 7620 \ \ \ \ \ \ \ \ 1117644

164 \ \ \ \ 16 \ \ \ \ \ \ \ \ \ \ \ 7620 \ \ \ \ \ \ \ \ 1125264

165 \ \ \ \ 16 \ \ \ \ \ \ \ \ \ \ \ 5316 \ \ \ \ \ \ \ \ 1130580

166 \ \ \ \ 16 \ \ \ \ \ \ \ \ \ \ \ 5316 \ \ \ \ \ \ \ \ 1135896

167 \ \ \ \ 16 \ \ \ \ \ \ \ \ \ \ \ 7620 \ \ \ \ \ \ \ \ 1143516

168 \ \ \ \ 16 \ \ \ \ \ \ \ \ \ \ \ 7620 \ \ \ \ \ \ \ \ 1151136

169 \ \ \ \ 16 \ \ \ \ \ \ \ \ \ \ \ 7620 \ \ \ \ \ \ \ \ 1158756

170 \ \ \ \ 16 \ \ \ \ \ \ \ \ \ \ \ 7620 \ \ \ \ \ \ \ \ 1166376

171 \ \ \ \ 16 \ \ \ \ \ \ \ \ \ \ \ 5316 \ \ \ \ \ \ \ \ 1171692

172 \ \ \ \ 16 \ \ \ \ \ \ \ \ \ \ \ 5316 \ \ \ \ \ \ \ \ 1177008

173 \ \ \ \ 16 \ \ \ \ \ \ \ \ \ \ \ 7620 \ \ \ \ \ \ \ \ 1184628

174 \ \ \ \ 16 \ \ \ \ \ \ \ \ \ \ \ 7620 \ \ \ \ \ \ \ \ 1192248

175 \ \ \ \ 16 \ \ \ \ \ \ \ \ \ \ \ 7620 \ \ \ \ \ \ \ \ 1199868

176 \ \ \ \ 16 \ \ \ \ \ \ \ \ \ \ \ 7620 \ \ \ \ \ \ \ \ 1207488

177 \ \ \ \ 16 \ \ \ \ \ \ \ \ \ \ \ 5316 \ \ \ \ \ \ \ \ 1212804

178 \ \ \ \ 16 \ \ \ \ \ \ \ \ \ \ \ 5316 \ \ \ \ \ \ \ \ 1218120

179 \ \ \ \ 16 \ \ \ \ \ \ \ \ \ \ \ 7620 \ \ \ \ \ \ \ \ 1225740

180 \ \ \ \ 16 \ \ \ \ \ \ \ \ \ \ \ 7620 \ \ \ \ \ \ \ \ 1233360

181 \ \ \ \ 16 \ \ \ \ \ \ \ \ \ \ \ 5316 \ \ \ \ \ \ \ \ 1238676

182 \ \ \ \ 16 \ \ \ \ \ \ \ \ \ \ \ 7620 \ \ \ \ \ \ \ \ 1246296

183 \ \ \ \ 16 \ \ \ \ \ \ \ \ \ \ \ 7620 \ \ \ \ \ \ \ \ 1253916

184 \ \ \ \ 16 \ \ \ \ \ \ \ \ \ \ \ 7620 \ \ \ \ \ \ \ \ 1261536

185 \ \ \ \ 16 \ \ \ \ \ \ \ \ \ \ \ 7620 \ \ \ \ \ \ \ \ 1269156

186 \ \ \ \ 16 \ \ \ \ \ \ \ \ \ \ \ 5316 \ \ \ \ \ \ \ \ 1274472

187 \ \ \ \ 16 \ \ \ \ \ \ \ \ \ \ \ 5316 \ \ \ \ \ \ \ \ 1279788

188 \ \ \ \ 16 \ \ \ \ \ \ \ \ \ \ \ 7620 \ \ \ \ \ \ \ \ 1287408

189 \ \ \ \ 16 \ \ \ \ \ \ \ \ \ \ \ 7620 \ \ \ \ \ \ \ \ 1295028

190 \ \ \ \ 16 \ \ \ \ \ \ \ \ \ \ \ 7620 \ \ \ \ \ \ \ \ 1302648

191 \ \ \ \ 16 \ \ \ \ \ \ \ \ \ \ \ 7620 \ \ \ \ \ \ \ \ 1310268

192 \ \ \ \ 16 \ \ \ \ \ \ \ \ \ \ \ 5316 \ \ \ \ \ \ \ \ 1315584

193 \ \ \ \ 16 \ \ \ \ \ \ \ \ \ \ \ 5316 \ \ \ \ \ \ \ \ 1320900

194 \ \ \ \ 16 \ \ \ \ \ \ \ \ \ \ \ 7620 \ \ \ \ \ \ \ \ 1328520

195 \ \ \ \ 16 \ \ \ \ \ \ \ \ \ \ \ 7620 \ \ \ \ \ \ \ \ 1336140

196 \ \ \ \ 16 \ \ \ \ \ \ \ \ \ \ \ 7620 \ \ \ \ \ \ \ \ 1343760

197 \ \ \ \ 16 \ \ \ \ \ \ \ \ \ \ \ 7620 \ \ \ \ \ \ \ \ 1351380

198 \ \ \ \ 16 \ \ \ \ \ \ \ \ \ \ \ 5316 \ \ \ \ \ \ \ \ 1356696

199 \ \ \ \ 16 \ \ \ \ \ \ \ \ \ \ \ 5316 \ \ \ \ \ \ \ \ 1362012

200 \ \ \ \ 16 \ \ \ \ \ \ \ \ \ \ \ 7620 \ \ \ \ \ \ \ \ 1369632

201 \ \ \ \ 16 \ \ \ \ \ \ \ \ \ \ \ 7620 \ \ \ \ \ \ \ \ 1377252

202 \ \ \ \ 16 \ \ \ \ \ \ \ \ \ \ \ 7620 \ \ \ \ \ \ \ \ 1384872

203 \ \ \ \ 16 \ \ \ \ \ \ \ \ \ \ \ 7620 \ \ \ \ \ \ \ \ 1392492

204 \ \ \ \ 16 \ \ \ \ \ \ \ \ \ \ \ 5316 \ \ \ \ \ \ \ \ 1397808

205 \ \ \ \ 16 \ \ \ \ \ \ \ \ \ \ \ 5316 \ \ \ \ \ \ \ \ 1403124

206 \ \ \ \ 16 \ \ \ \ \ \ \ \ \ \ \ 7620 \ \ \ \ \ \ \ \ 1410744

207 \ \ \ \ 16 \ \ \ \ \ \ \ \ \ \ \ 7620 \ \ \ \ \ \ \ \ 1418364

208 \ \ \ \ 16 \ \ \ \ \ \ \ \ \ \ \ 7620 \ \ \ \ \ \ \ \ 1425984

209 \ \ \ \ 16 \ \ \ \ \ \ \ \ \ \ \ 7620 \ \ \ \ \ \ \ \ 1433604

210 \ \ \ \ 16 \ \ \ \ \ \ \ \ \ \ \ 5316 \ \ \ \ \ \ \ \ 1438920

211 \ \ \ \ 16 \ \ \ \ \ \ \ \ \ \ \ 5316 \ \ \ \ \ \ \ \ 1444236

212 \ \ \ \ 16 \ \ \ \ \ \ \ \ \ \ \ 7620 \ \ \ \ \ \ \ \ 1451856

213 \ \ \ \ 16 \ \ \ \ \ \ \ \ \ \ \ 7620 \ \ \ \ \ \ \ \ 1459476

214 \ \ \ \ 16 \ \ \ \ \ \ \ \ \ \ \ 7620 \ \ \ \ \ \ \ \ 1467096

215 \ \ \ \ 16 \ \ \ \ \ \ \ \ \ \ \ 7620 \ \ \ \ \ \ \ \ 1474716

216 \ \ \ \ 16 \ \ \ \ \ \ \ \ \ \ \ 5316 \ \ \ \ \ \ \ \ 1480032

217 \ \ \ \ 16 \ \ \ \ \ \ \ \ \ \ \ 7620 \ \ \ \ \ \ \ \ 1487652

218 \ \ \ \ 16 \ \ \ \ \ \ \ \ \ \ \ 7620 \ \ \ \ \ \ \ \ 1495272

219 \ \ \ \ 16 \ \ \ \ \ \ \ \ \ \ \ 5316 \ \ \ \ \ \ \ \ 1500588

220 \ \ \ \ 16 \ \ \ \ \ \ \ \ \ \ \ 5316 \ \ \ \ \ \ \ \ 1505904

221 \ \ \ \ 16 \ \ \ \ \ \ \ \ \ \ \ 7620 \ \ \ \ \ \ \ \ 1513524

222 \ \ \ \ 16 \ \ \ \ \ \ \ \ \ \ \ 7620 \ \ \ \ \ \ \ \ 1521144

223 \ \ \ \ 16 \ \ \ \ \ \ \ \ \ \ \ 7620 \ \ \ \ \ \ \ \ 1528764

224 \ \ \ \ 16 \ \ \ \ \ \ \ \ \ \ \ 7620 \ \ \ \ \ \ \ \ 1536384

225 \ \ \ \ 16 \ \ \ \ \ \ \ \ \ \ \ 5316 \ \ \ \ \ \ \ \ 1541700

226 \ \ \ \ 16 \ \ \ \ \ \ \ \ \ \ \ 5316 \ \ \ \ \ \ \ \ 1547016

227 \ \ \ \ 16 \ \ \ \ \ \ \ \ \ \ \ 7620 \ \ \ \ \ \ \ \ 1554636

228 \ \ \ \ 16 \ \ \ \ \ \ \ \ \ \ \ 7620 \ \ \ \ \ \ \ \ 1562256

229 \ \ \ \ 16 \ \ \ \ \ \ \ \ \ \ \ 7620 \ \ \ \ \ \ \ \ 1569876

230 \ \ \ \ 16 \ \ \ \ \ \ \ \ \ \ \ 7620 \ \ \ \ \ \ \ \ 1577496

231 \ \ \ \ 16 \ \ \ \ \ \ \ \ \ \ \ 5316 \ \ \ \ \ \ \ \ 1582812

232 \ \ \ \ 16 \ \ \ \ \ \ \ \ \ \ \ 5316 \ \ \ \ \ \ \ \ 1588128

233 \ \ \ \ 16 \ \ \ \ \ \ \ \ \ \ \ 7620 \ \ \ \ \ \ \ \ 1595748

234 \ \ \ \ 16 \ \ \ \ \ \ \ \ \ \ \ 7620 \ \ \ \ \ \ \ \ 1603368

235 \ \ \ \ 16 \ \ \ \ \ \ \ \ \ \ \ 7620 \ \ \ \ \ \ \ \ 1610988

236 \ \ \ \ 16 \ \ \ \ \ \ \ \ \ \ \ 7620 \ \ \ \ \ \ \ \ 1618608

237 \ \ \ \ 16 \ \ \ \ \ \ \ \ \ \ \ 5316 \ \ \ \ \ \ \ \ 1623924

238 \ \ \ \ 16 \ \ \ \ \ \ \ \ \ \ \ 5316 \ \ \ \ \ \ \ \ 1629240

239 \ \ \ \ 16 \ \ \ \ \ \ \ \ \ \ \ 7620 \ \ \ \ \ \ \ \ 1636860

240 \ \ \ \ 16 \ \ \ \ \ \ \ \ \ \ \ 7620 \ \ \ \ \ \ \ \ 1644480

241 \ \ \ \ 16 \ \ \ \ \ \ \ \ \ \ \ 7620 \ \ \ \ \ \ \ \ 1652100

242 \ \ \ \ 16 \ \ \ \ \ \ \ \ \ \ \ 7620 \ \ \ \ \ \ \ \ 1659720

243 \ \ \ \ 16 \ \ \ \ \ \ \ \ \ \ \ 5316 \ \ \ \ \ \ \ \ 1665036

244 \ \ \ \ 16 \ \ \ \ \ \ \ \ \ \ \ 5316 \ \ \ \ \ \ \ \ 1670352

245 \ \ \ \ 16 \ \ \ \ \ \ \ \ \ \ \ 7620 \ \ \ \ \ \ \ \ 1677972

246 \ \ \ \ 16 \ \ \ \ \ \ \ \ \ \ \ 7620 \ \ \ \ \ \ \ \ 1685592

247 \ \ \ \ 16 \ \ \ \ \ \ \ \ \ \ \ 7620 \ \ \ \ \ \ \ \ 1693212

248 \ \ \ \ 16 \ \ \ \ \ \ \ \ \ \ \ 7620 \ \ \ \ \ \ \ \ 1700832

249 \ \ \ \ 16 \ \ \ \ \ \ \ \ \ \ \ 5316 \ \ \ \ \ \ \ \ 1706148

250 \ \ \ \ 16 \ \ \ \ \ \ \ \ \ \ \ 5316 \ \ \ \ \ \ \ \ 1711464

251 \ \ \ \ 16 \ \ \ \ \ \ \ \ \ \ \ 7620 \ \ \ \ \ \ \ \ 1719084

252 \ \ \ \ 16 \ \ \ \ \ \ \ \ \ \ \ 7620 \ \ \ \ \ \ \ \ 1726704

253 \ \ \ \ 16 \ \ \ \ \ \ \ \ \ \ \ 5316 \ \ \ \ \ \ \ \ 1732020

254 \ \ \ \ 16 \ \ \ \ \ \ \ \ \ \ \ 7620 \ \ \ \ \ \ \ \ 1739640

255 \ \ \ \ 16 \ \ \ \ \ \ \ \ \ \ \ 7620 \ \ \ \ \ \ \ \ 1747260

256 \ \ \ \ 16 \ \ \ \ \ \ \ \ \ \ \ 7620 \ \ \ \ \ \ \ \ 1754880

257 \ \ \ \ 16 \ \ \ \ \ \ \ \ \ \ \ 7620 \ \ \ \ \ \ \ \ 1762500

258 \ \ \ \ 16 \ \ \ \ \ \ \ \ \ \ \ 5316 \ \ \ \ \ \ \ \ 1767816

259 \ \ \ \ 16 \ \ \ \ \ \ \ \ \ \ \ 5316 \ \ \ \ \ \ \ \ 1773132

260 \ \ \ \ 16 \ \ \ \ \ \ \ \ \ \ \ 7620 \ \ \ \ \ \ \ \ 1780752

261 \ \ \ \ 16 \ \ \ \ \ \ \ \ \ \ \ 7620 \ \ \ \ \ \ \ \ 1788372

262 \ \ \ \ 16 \ \ \ \ \ \ \ \ \ \ \ 7620 \ \ \ \ \ \ \ \ 1795992

263 \ \ \ \ 16 \ \ \ \ \ \ \ \ \ \ \ 7620 \ \ \ \ \ \ \ \ 1803612

264 \ \ \ \ 16 \ \ \ \ \ \ \ \ \ \ \ 5316 \ \ \ \ \ \ \ \ 1808928

265 \ \ \ \ 16 \ \ \ \ \ \ \ \ \ \ \ 5316 \ \ \ \ \ \ \ \ 1814244

266 \ \ \ \ 16 \ \ \ \ \ \ \ \ \ \ \ 7620 \ \ \ \ \ \ \ \ 1821864

267 \ \ \ \ 16 \ \ \ \ \ \ \ \ \ \ \ 7620 \ \ \ \ \ \ \ \ 1829484

268 \ \ \ \ 16 \ \ \ \ \ \ \ \ \ \ \ 7620 \ \ \ \ \ \ \ \ 1837104

269 \ \ \ \ 16 \ \ \ \ \ \ \ \ \ \ \ 7620 \ \ \ \ \ \ \ \ 1844724

270 \ \ \ \ 16 \ \ \ \ \ \ \ \ \ \ \ 5316 \ \ \ \ \ \ \ \ 1850040

271 \ \ \ \ 16 \ \ \ \ \ \ \ \ \ \ \ 5316 \ \ \ \ \ \ \ \ 1855356

272 \ \ \ \ 16 \ \ \ \ \ \ \ \ \ \ \ 7620 \ \ \ \ \ \ \ \ 1862976

273 \ \ \ \ 16 \ \ \ \ \ \ \ \ \ \ \ 7620 \ \ \ \ \ \ \ \ 1870596

274 \ \ \ \ 16 \ \ \ \ \ \ \ \ \ \ \ 7620 \ \ \ \ \ \ \ \ 1878216

275 \ \ \ \ 16 \ \ \ \ \ \ \ \ \ \ \ 7620 \ \ \ \ \ \ \ \ 1885836

276 \ \ \ \ 16 \ \ \ \ \ \ \ \ \ \ \ 5316 \ \ \ \ \ \ \ \ 1891152

277 \ \ \ \ 16 \ \ \ \ \ \ \ \ \ \ \ 5316 \ \ \ \ \ \ \ \ 1896468

278 \ \ \ \ 16 \ \ \ \ \ \ \ \ \ \ \ 7620 \ \ \ \ \ \ \ \ 1904088

279 \ \ \ \ 16 \ \ \ \ \ \ \ \ \ \ \ 7620 \ \ \ \ \ \ \ \ 1911708

280 \ \ \ \ 16 \ \ \ \ \ \ \ \ \ \ \ 7620 \ \ \ \ \ \ \ \ 1919328

281 \ \ \ \ 16 \ \ \ \ \ \ \ \ \ \ \ 7620 \ \ \ \ \ \ \ \ 1926948

282 \ \ \ \ 16 \ \ \ \ \ \ \ \ \ \ \ 5316 \ \ \ \ \ \ \ \ 1932264

283 \ \ \ \ 16 \ \ \ \ \ \ \ \ \ \ \ 5316 \ \ \ \ \ \ \ \ 1937580

284 \ \ \ \ 16 \ \ \ \ \ \ \ \ \ \ \ 7620 \ \ \ \ \ \ \ \ 1945200

285 \ \ \ \ 16 \ \ \ \ \ \ \ \ \ \ \ 7620 \ \ \ \ \ \ \ \ 1952820

286 \ \ \ \ 16 \ \ \ \ \ \ \ \ \ \ \ 7620 \ \ \ \ \ \ \ \ 1960440

287 \ \ \ \ 16 \ \ \ \ \ \ \ \ \ \ \ 7620 \ \ \ \ \ \ \ \ 1968060

288 \ \ \ \ 16 \ \ \ \ \ \ \ \ \ \ \ 5316 \ \ \ \ \ \ \ \ 1973376

Found 288 coverings in 0 secs, 20000 usecs

Average Estimation on 288 paths:

\ \ \ \ \ \ \ \ \ \ \ \ \ \ \ \ \ \ \ \ \ \ \ \ \ \ \ \ \ \ \ \ \ \ \ \ \ \ \
\ \ \ \ \ \ 6852}

The first column prints the solution's number, the second the depth reached
(always 16 for size-2 grid), the third one the estimation of the tree size,
and the last one the cumulated estimation used to provide the average on the
last line. \ Here, in average, 6852 nodes are explored in the backtrack tree
problem space.

\paragraph{Further results.}The Dancing Links implementation makes it very
easy to experiment with several variations of the Su-Doku grids and puzzles.
Adding further constraints to the problem is simply a matter of adding columns
to the A matrix used by the algorithm. \ In the diagonal variant, for
instance, where a Su-Doku grid is considered valid if, in addition, all
numbers in both diagonals are also different, $2 n$ columns are added to the A
matrix to account for the $n$ possible positions of each $1 \ldots n^2$ figure
in each diagonal.

In this variant, running the Dancing Links for enumeration yields the 48
unique solutions for a size 2 diagonal Su-Doku problem (a 4-by-4 grid) and an
average of 3666 nodes explored in the backtrack tree problem space. \ Note
that the exploration space/time complexity is roughly halved on this instance.

On a different track, the Dancing Links algorithm was successfully used for
experimenting with the Langford problems which combinatorial nature,
ultimately relying on permutation constraints, lends it perfectly to Dancing
Links-based study.

\end{document}